\documentclass[10pt,twocolumn,letterpaper]{article}

\usepackage[pagenumbers]{cvpr} 
\usepackage{colortbl}
\usepackage{float}
\usepackage{listings}
\usepackage{comment}
\usepackage{pifont}
\usepackage{booktabs}
\usepackage[table]{xcolor}
\usepackage{array} 
\usepackage{booktabs}
\usepackage{adjustbox}
\usepackage{multirow}
\usepackage[para]{footmisc}
\definecolor{pearDark}{RGB}{0,0,128} 
\definecolor{mycolor_green}{RGB}{0,128,0} 
\definecolor{cvprblue}{rgb}{0.21,0.49,0.74}
\usepackage[pagebackref,breaklinks,colorlinks,allcolors=cvprblue]{hyperref}

\def\paperID{} 
\def\confName{CVPR}
\def\confYear{2026}

\title{Design Your Ad: Personalized Advertising Image and Text Generation with Unified Autoregressive Models}

\author{
	Yexing Xu$^{1*}$\quad Wei Feng$^{2\dagger}$\quad Shen Zhang$^{2}$\quad Haohan Wang$^{2}$\quad Yuxin Qin$^{2}$\quad Yaoyu Li$^{2}$\quad Ao Ma$^{2}$\\
    Yuhao Luo$^{2}$\quad Lu Wang$^{2}$\quad Xudong Ren$^{2}$\quad Haoran Wang$^{2}$\quad Run Ling$^{3}$\quad Zheng Zhang$^{2}$\quad Jingjing Lv$^{2}$\\
    Junjie Shen$^{2}$\quad Ching Law$^{2}$\quad Longguang Wang$^{1}$\quad Yulan Guo$^{1\ddagger}$ \\
	\vspace{-0.8em} \\
	{ $^1$Sun Yat-Sen University \quad $^2$JD.COM \quad $^3$Northeastern University}\\
}
\begin{document}
\maketitle
\renewcommand{\thefootnote}{\fnsymbol{footnote}}
\footnotetext[1]{ Work done while interning at JD.COM \quad $^\ddagger$ Corresponding author. \\
$^\dagger$  The first two authors contributed equally to this research.
}
\begin{abstract}
Generating realistic and user-preferred advertisements is a key challenge in e-commerce. Existing approaches utilize multiple independent models driven by click-through-rate (CTR) to controllably create attractive image or text advertisements. However, their pipelines lack cross-modal perception and rely on CTR that only reflects average preferences. Therefore, we explore jointly generating personalized image-text advertisements from historical click behaviors. We first design a Unified Advertisement Generative model (Uni-AdGen) that employs a single autoregressive framework to produce both advertising images and texts. By incorporating a foreground perception module and instruction tuning, Uni-AdGen enhances the realism of the generated content. To further personalize advertisements, we equip Uni-AdGen with a coarse-to-fine preference understanding module that effectively captures user interests from noisy multimodal historical behaviors to drive personalized generation. Additionally, we construct the first large-scale Personalized Advertising image-text dataset (PAd1M) and introduce a Product Background Similarity (PBS) metric to facilitate training and evaluation. Extensive experiments show that our method outperforms baselines in general and personalized advertisement generation. Our project is available at https://github.com/JD-GenX/Uni-AdGen.

\end{abstract}    
\section{Introduction}
\label{sec:intro}

\begin{figure*}[ht]
    \centering
    \includegraphics[width=\linewidth]{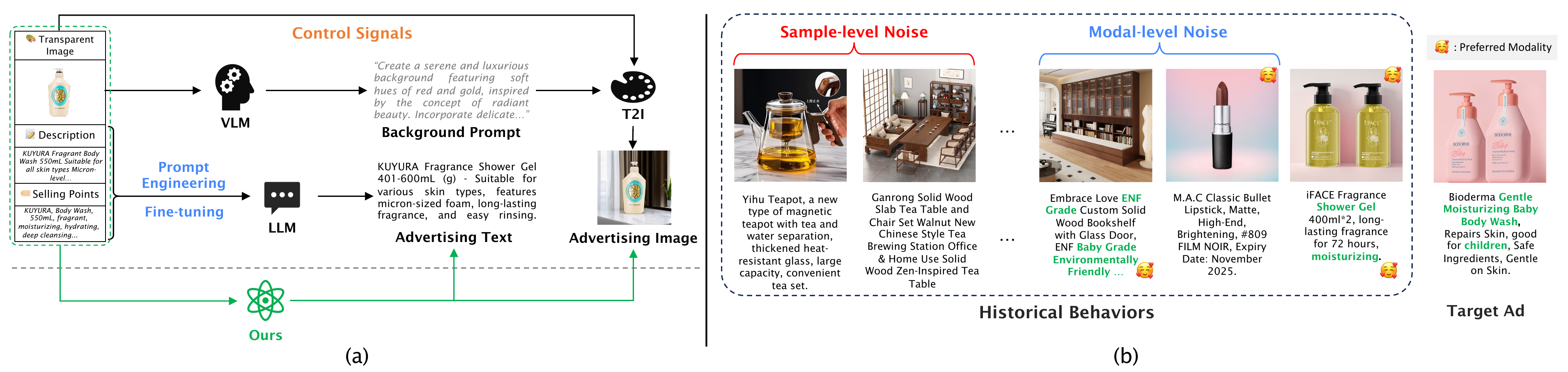}
    \caption{Problems in existing methods and user historical behaviors. (a) Existing methods use separate models for advertisement generation, making systems complex and modalities isolated. Our approach uses a unified autoregressive model to generate advertising images and texts jointly. (b) Sample-level noise refers to historical products that are visually or textually unrelated to the target product, such as the teapot and table in the figure. Modal-level noise reflects users' dynamic preference across modalities. For example, the user may prefer the keyword ``baby grade" in bookshelf, while favoring a pink background image of the lipstick.}
    \label{fig:problems}
\end{figure*}

E-commerce platforms typically use a combination of advertising images and texts to promote products collaboratively, which is important for enhancing shopping experiences. However, manually designing these advertising materials is costly and inefficient. With the rapid advancement of generative models \cite{gpt4o,flux,qwen3}, research focus has increasingly shifted toward generative solutions for advertising images and texts.

To generate realistic advertising content, existing approaches \cite{cxy, reliablead, postermaker, chen2025t, chen2025ctr, cao2024product2img, lfh} mainly combine multiple independent models illustrated in Fig.~\ref{fig:problems}(a): a text-to-image (T2I) model \cite{dalle2,betker2023improving,wang2023imagen} first generates the advertising images from a background prompt produced by a Visual Language Model (VLM) \cite{qwen2_5_vl, Janus, Janus-Pro} based on the product transparent image, while a Large Language Model (LLM) \cite{gpt4o,deepseekr1,qwen2_5,qwen3} creates advertising texts based on the product description and selling points. In this process, feature injection \cite{controlar, controlnet, lfh}, and inpainting \cite{postermaker, reliablead, chen2025t, cao2024product2img} are employed to produce control signals for image generation. Fine-tuning \cite{wei2022creater, li2022culg, shao2021controllable, jin2023towards} and prompt engineering \cite{autoprompt, yu2021attribute, gpt2} are adopted to alleviate hallucinations in generated texts. However, these methods process images and texts separately, increasing system complexity and limiting effective collaboration between the two modalities. To address this issue, we propose a Unified Advertisement Generative model (Uni-AdGen) to incorporate images and texts into a single generation process in an autoregressive framework. This framework discretizes multimodal inputs into tokens and generates content through next-token prediction, offering architectural simplicity while enabling natural cross-modal interaction. Additionally, Uni-AdGen incorporates designed controls for image and text generation. The image part utilizes a foreground perception module to extract structural guidance from the product's transparent image, while the text part utilizes instruction tuning to align generated content with provided selling points. Together, these control approaches help Uni-AdGen improve the visual and factual consistency in generated images and texts. 

Building on general advertisement generation, recent studies \cite{chen2025ctr, cxy, wang2025beyond} have shifted toward personalized generation by incorporating click-through rates (CTR) to enhance the appeal of generated advertisements. However, CTR signals primarily reflect average preferences and are often insufficient for producing advertisements that align with individual user interests. Therefore, we explore generating more tailored image-text advertisements based on users' historical behaviors. The key challenge lies in the sample-level and modal-level noise in user behaviors. As shown in Fig.~\ref{fig:problems}(b), sample-level noise includes irrelevant products (e.g., the teapot and table), whose visual and textual style poorly match the target product. Modal-level noise arises when the modality driving user clicks varies, for example the user may be interested in either pink background image or keywords like ``moisturizing". Therefore, relying solely on one modality for preference extraction leads to incomplete and biased user modeling. In this paper, we design a coarse-to-fine preference understanding module in Uni-AdGen that jointly utilizes images and texts to extract cross-modal user interests. In the coarse stage, it first selects the top-N relevant historical clicks based on product similarity, reducing sample-level noise. Then, it performs fine-grained denoising by suppressing irrelevant tokens to address modal-level noise. This module enables Uni-AdGen to generate more precise personalized advertisements.

To support personalized advertisement generation research, we construct the first large-scale \textbf{P}ersonalized \textbf{Ad}vertising image-text dataset (PAd1M). It includes millions of users and 8.5 million products, significantly expanding existing datasets in scale. We also propose Product Background Similarity (PBS), a novel evaluation metric that effectively ignores interference from the same foreground products, enabling more accurate style similarity measurement for personalized advertisement generation.

The main contributions are summarized as follows:

\begin{itemize} 
    \item To the best of our knowledge, we propose Uni-AdGen, the first unified autoregressive model for joint and controllable image-text advertisement generation.
    \item We design a coarse-to-fine preference understanding module to effectively leverage noisy, multi-modal user behaviors for personalized advertising generation.
    \item We introduce the Personalized Advertising Image-text dataset (PAd1M) and a Product Background Similarity (PBS) metric to benchmark personalized advertisement generation. Extensive experiments show that our method achieves state-of-the-art performance in both general and personalized settings.
\end{itemize}

\begin{figure*}[ht]
    \centering
    \includegraphics[width=\linewidth]{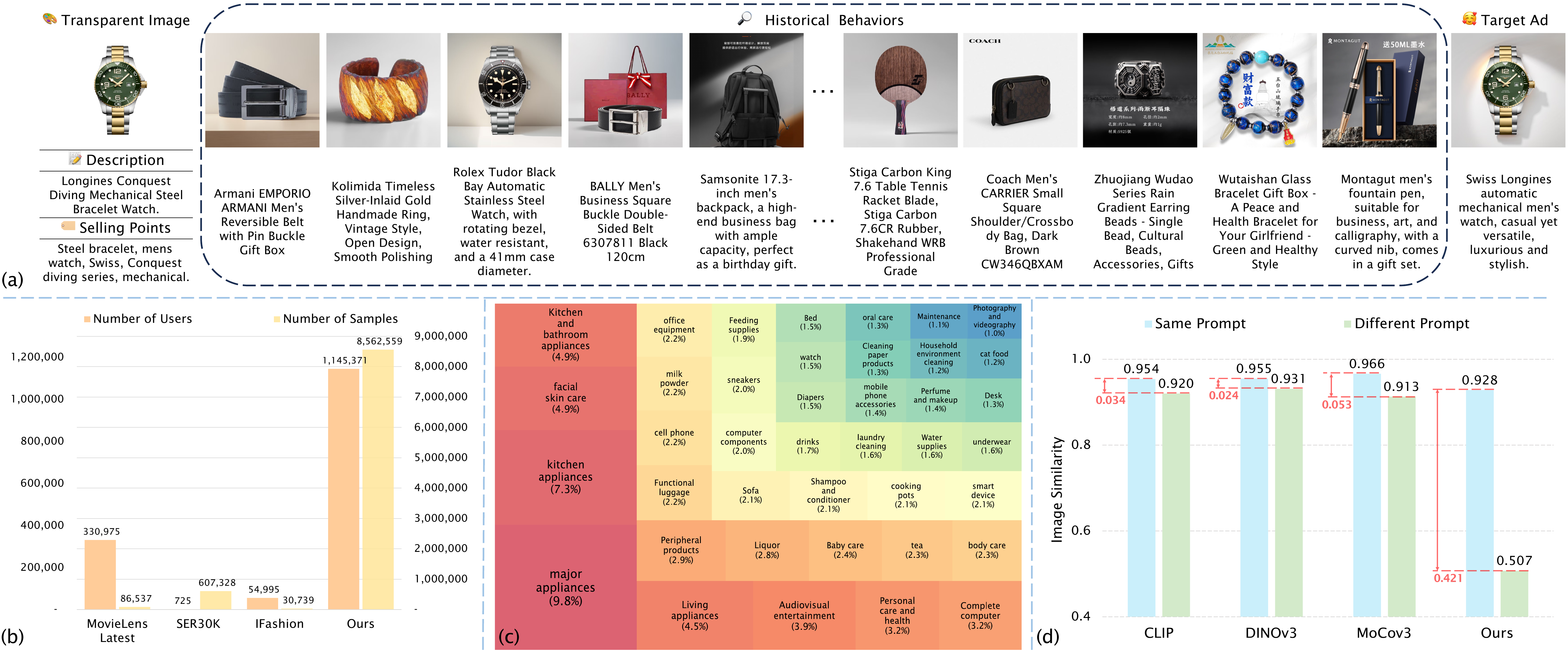}
    \caption{An overview of PAd1M and experimental results of the PBS metric. (a) Visualization of PAd1M, showing the target advertisement together with users’ historical behaviors and associated product information; (b) Scale comparison of personalized datasets, where PAd1M contains substantially more users and samples than existing datasets; (c) Treemap of the top 40 product categories in PAd1M with rectangle area proportional to sample count, highlighting diverse types of products; (d) Results on image-similarity evaluation metrics. Our PBS metric can better distinguish images with different backgrounds.}
    \label{fig:dataset}
\end{figure*}

\section{Related Work}
\label{sec:related_work}

\subsection{Advertisement Generation}
The objective of advertising text and image generation is to craft compelling content that enhances users’ interest in products. With the development of LLMs~\cite{gpt4o,deepseekr1,qwen2_5,qwen3} and Diffusion Models~\cite{dalle2,betker2023improving,wang2023imagen,rombach2022high,podell2023sdxl,flux,esser2024scaling,controlnet}, AIGC-based advertisements are occupying an increasingly significant position in advertising delivery. 
Specifically, for advertising texts, most works aim at improving the intrinsic text quality, especially in fluency~\cite{li2022culg,wei2022creater}, authenticity~\cite{chen2025enhancing,shao2021controllable} and diversity~\cite{shao2021controllable,jin2023towards} of the contents.
Meanwhile for advertising image, existing works mainly focus on the controllable generation conditioned on additional advertising-related information, such as user-defined~\cite{postermaker,fan2025autopp,li2023relation} and automatically generated~\cite{lfh,chen2025t,lin2023autoposter,lu2025uni} layout, reference image with specific styles~\cite{whh,zhao2025dreampainter}, category-wise prior knowledge~\cite{whh,cao2024product2img}, or their combination~\cite{chen2024enhancing,sun2025minimal}. 
Nonetheless, the enhancements in content quality cannot guarantee a corresponding increase in advertising effectiveness.

    Beyond the offline metrics, the direct optimization towards the user preferences related to online clicks is attracting growing interest. Earlier methods for optimizing CTR in advertising text mostly resemble BART fine-tuning~\cite{kanungo2022cobart} or contrastive learning~\cite{wei2022creater}. Recently, such online metrics have been advanced by the rapidly evolving reinforcement learning algorithms~\cite{jiang2025improving,chen2025ctr,wang2025beyond,chen2025hllm} and in-context learning~\cite{murakami2025adparaphrase}.
Such progress is relatively slower in the field of image generation, with only Chen~\textit{et al.}~\cite{cxy}, Yang~\textit{et al.}~\cite{shopeectr}, and Lu~\textit{et al.}~\cite{lu2026one} recently beginning to align click preference directly in image generation.
These efforts, however, derive preference reward signals from the aggregate CTR of all users. This actually assumes a general and consistent preference on advertisement and overlooks the complexity and diversity of individual preferences. Hence such approaches often lead to the suboptimal performance unavoidably.

\subsection{Autoregressive Generation}

The autoregressive generation, where the content is produced by a next-token prediction mechanism, has dominated natural language processing due to the remarkable breakthrough achieved by LLMs. 
The autoregressive models could encode various requirements in content or format into the input sequences naturally, and generate proper texts through in-context learning~\cite{murakami2025adparaphrase} and fine-tuning~\cite{jiang2025improving,chen2025ctr,wang2025beyond,chen2025hllm}, owing to the robust generalization capabilities of pretrained LLMs~\cite{gpt4o,deepseekr1,qwen2_5,qwen3}.
Such a flexible characteristic enables autoregressive models to be widely applied in various downstream tasks.

In contrast, autoregressive models used not to be the first choice for image generation, as they historically underperformed aesthetically compared to diffusion models. This is recently alleviated with scaling up the model and data during training~\cite{Janus-Pro,xin2025lumina}. 
Meanwhile, the advancement in controllable image generation is making it possible for autoregressive models to be applied as flexibly as diffusion models. 
Concretely, the pre-filling strategy is mostly leveraged where the extra condition image is encoded and integrated into the input sequence~\cite{li2024controlvar,xin2025lumina,chen2025context,han2025controlthinker}. An alternative is to intervene in the decoding process during the next-token prediction~\cite{yao2024car,xu2025scalar,controlar}, which is similar to the mechanism used in diffusion models~\cite{controlnet}.

Furthermore, with a similar generation mechanism, the autoregressive paradigm exhibits potential for unifying image and text generation within the same model architecture. This inspires us to generate closely connected advertising images and texts simultaneously, thereby preventing the information redundancy or conflict arising from the isolated generation of advertisement components.

\section{Dataset}
\label{sec:dataset}

\subsection{Dataset Construction}
To advance studies in generating personalized advertisements, we construct a large-scale and diverse Personalized Advertising image-text (PAd1M) dataset through a systematic multi-stage pipeline. We collect advertising images and texts from JD.com with a large number of active users and products. Each product typically provides more than ten candidate images and texts, ensuring that the diverse preferences can be fully detected. To enable reliable preference modeling, we collect complete user click histories over both images and texts, filtering out users with insufficient activity to reduce noise. This yields a dataset of 1,145,371 users, with 18,923,555 clicked product images and texts, averaging more than sixteen multimodal historical behaviors per user. 
Subsequently, for each user, we randomly select one clicked image–text record as the target and extract the product foreground from the target image using Grounded SAM~\cite{groundedsam}. We then obtain the seller-provided product description and selling points to construct our PAd1M dataset. 
As shown in Fig.~\ref{fig:dataset}(a), each user includes the target image–text pair with the transparent image, as well as the product description and selling points, and is also associated with abundant historical images and texts that reflect prior interests and enable fine-grained preference modeling.

\subsection{Dataset Characteristics}
Compared to existing publicly available datasets, the dataset we propose offers three key advantages:

\begin{itemize}
\item \noindent \textbf{Large-Scale:} Most existing personalized datasets~\citep{ser30k, movielens, ifashion} are restricted in both user and sample numbers. PAd1M spans millions of users and product samples as shown in Fig.~\ref{fig:dataset}(b), substantially exceeding prior datasets and enabling practically relevant large‑scale research on personalized preferences.

\item \noindent \textbf{Comprehensive Coverage:} We visualize the top 40 product categories in PAd1M using a treemap as shown in Fig.~\ref{fig:dataset}(c). PAd1M encompasses a wide range of advertised product categories far beyond the scope of single-domain datasets~\citep{ser30k, movielens, ifashion}. Meanwhile, by including both advertising images and texts, it provides comprehensive multimodal data that supports richer and more diverse modeling of user preferences.

\item \noindent \textbf{Field Innovation:} Distinct from previous datasets~\citep{chakraborty2024maxmin, ramesh2024group} that primarily focus on CTR, which reflects group-level preferences, PAd1M is the first dataset in the advertisement domain to explicitly establish preference links at the individual user. This enables scalable and fine-grained analysis of heterogeneous user preferences, advancing personalized advertisement research to the individual level.
\end{itemize}

\subsection{Evaluation Metrics}

In the context of personalized advertisement generation, the fundamental principle for evaluating model performance is to measure the similarity between generated advertisements and those that users actually engage with. Specifically, we compute both text and image similarity between the target and generated advertisements. For text similarity, we adopt standard metrics widely used in natural language generation, including BLEU~\cite{bleu} and ROUGE~\cite{rouge}, which measure n-gram overlap and the longest common subsequence between generated and target texts, respectively. For Chinese text processing, we apply character-level tokenization following standard practice. 

Image similarity evaluation metrics such as CLIP~\citep{openclip} and DINOv3~\citep{dinov3} prioritize foreground objects and often overlook the background context that is critical for personalized advertisements. Therefore, we propose Product Background Similarity (PBS), a metric based on MoCov3~\citep{mocov3} with an improved training strategy, trained on 681,123 pairs of images of the same product generated by multiple different models~\cite{podell2023sdxl,flux,cxy,shopeectr,postermaker,reliablead}. In each pair, the product foreground remains consistent while the background varies. For training, each image is treated as a positive sample with itself, while the paired image with a different background serves as a negative sample. This strategy emphasizes background differences, reduces foreground bias, and improves sensitivity to context. For evaluation, we extract features using the trained encoder and compute cosine similarity.

To validate the effectiveness of PBS in evaluating personalized advertising images where background context is important, we conducted a comparative study using two prompt sets from GPT-4o~\citep{gpt4o} and Qwen2.5-VL~\citep{qwen2_5_vl}, with 3,000 images generated by Flux-Fill~\citep{flux}, PosterMaker~\citep{postermaker}, and ReliableAd~\citep{reliablead}. As shown in Fig.~\ref{fig:dataset}(d), PBS attains a 0.421 similarity gap between same and different background images, whereas CLIP, DINOv3, and MoCov3 are all below 0.05, demonstrating that PBS is highly sensitive to background variations.

\begin{figure*}[ht]
    \centering
    \includegraphics[width=0.9\linewidth]{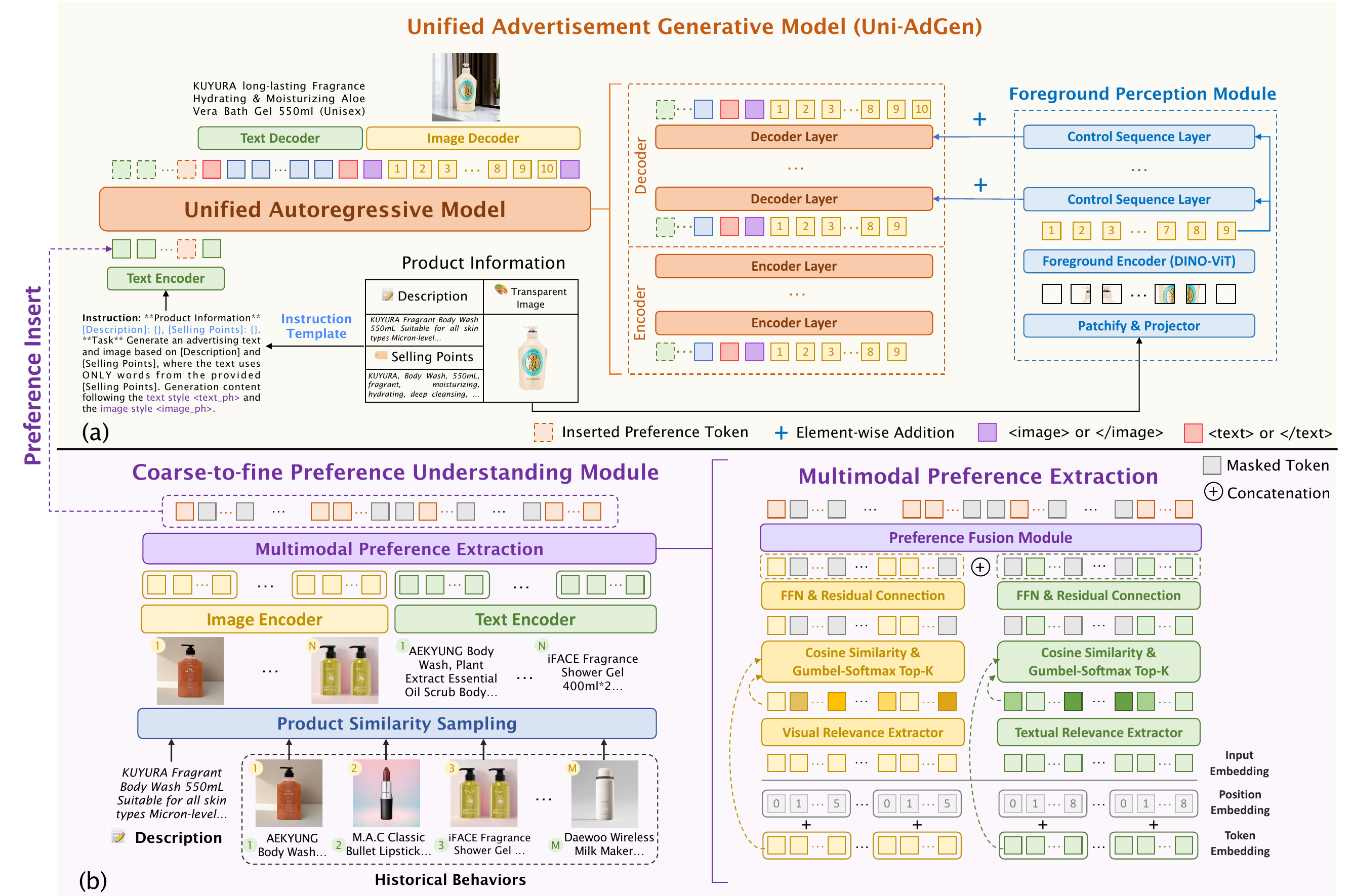}
    \caption{An overview of our method. Our framework combines (a) the unified advertisement generative model (Uni-AdGen) with (b) the coarse-to-fine preference understanding module. Uni-AdGen creates controllable advertising content from product information, while the preference understanding module personalizes the output by modeling user preferences from historical behaviors.}
    \label{fig:pipeline}
\end{figure*}

\section{Method}
\label{sec:method}
\subsection{Task Definition}
\label{sec:task_def}
We aim to model user preferences from their historical behaviors and leverage them to produce tailored advertising images and texts. Specifically, for each user, we record his or her historical click behaviors as a sequence of $L$ image-text pairs $\{\left(I_j, T_j\right)\}_{j=1}^{L}$. Based on these behaviors, we aim to use the target product information $P$ to generate an image-text advertisement $(I^{pred}, T^{pred})$ that closely aligns with the actual click results $(I^{GT}, T^{GT})$ of the user. The product information includes a transparent image $I^t$, a description $D$, and selling points $W$, which are typically provided by sellers on e-commerce platforms. In summary, the task can be defined as:
\begin{equation}
    \mathop{\text{argmax}}\limits_{\theta}\,\mathcal{S}\left((I^{GT}, T^{GT}), f_{\theta}\left(P, \{\left(I_j, T_j\right)\}_{j=1}^{L}\right)\right),
\end{equation}
where $f_{\theta}$ denotes the neural network for advertising image and text prediction, and $\mathcal{S}(\cdot, \cdot)$ is used to measure the similarity between inputs.

\subsection{Overview}
In Sec.~\ref{sec:general_ad_gen}, we first build a Unified Advertisement Generative model (Uni-AdGen) on an autoregressive framework combined with specialized controls for general advertisement generation. In Sec.~\ref{sec:personalized_ad_gen}, we further extend Uni-AdGen with a coarse-to-fine preference understanding module to support personalized advertisement generation, which learns precise user preferences from noisy multimodal historical behaviors.

\subsection{General Ad Generation}
\label{sec:general_ad_gen}

\subsubsection{Unified Advertisement Generation}
As shown in Fig.~\ref{fig:pipeline}(a), Uni-AdGen adopts an autoregressive vision-language architecture \cite{Janus, Janus-Pro} for image-text advertisement generation. 
The model takes a structured instruction with the task definition, product description, and selling points to guide generation. We use special tokens $\texttt{<text>}$ and $\texttt{</text>}$ to demarcate the text generation range, followed by the $\texttt{<image>}$ to trigger image generation and $\texttt{</image>}$ to mark its completion. The model outputs are connected to separate image and text decoders that decode the predicted tokens into their respective modalities. The image decoder employs a VQ-GAN \cite{vqgan} architecture, mapping discrete tokens to pixel space through codebook lookup. Thanks to the multimodal paradigm, Uni-AdGen achieves end-to-end advertisement generation without multiple independent models required by existing methods.

The training of Uni-AdGen follows the next-token prediction paradigm. Given the instruction 
sequence with $N$ tokens $\mathbf{s} = \{\mathbf{s_1}, \mathbf{s_2}, \dots, \mathbf{s_N}\}$, the model autoregressively predicts the text sequence with $M$ tokens $\mathbf{t} = \{\mathbf{t_1}, \mathbf{t_2}, \dots, \mathbf{t_M}\}$ and image sequence with $K$ tokens $\mathbf{g} = \{\mathbf{g_1}, \mathbf{g_2}, \dots, \mathbf{g_K}\}$. We jointly optimize both the advertising text and image generation tasks for training. The 
text generation loss maximizes the likelihood of the text token given the input sequence and the previously generated text tokens:
\begin{equation}
    \mathcal{L}_{text}=\sum_i\text{log}p_{\theta}\left(\mathbf{t_i} \mid \mathbf{s}, \mathbf{t_{1:i-1}}\right),
\end{equation}
where the $p_{\theta}$ denotes the condition probability of Uni-AdGen with learnable parameters $\theta$. The image generation loss maximizes the likelihood of the image token conditioned on the input sequence, text tokens, and the previously generated image tokens:
\begin{equation}
    \mathcal{L}_{img}=\sum_i\text{log}p_{\theta}\left(\mathbf{g_i} \mid \mathbf{s}, \mathbf{t},\mathbf{g_{1:i-1}}\right).
\end{equation}
The overall loss can be defined as:
\begin{equation}
    \mathcal{L}=\lambda_{text}\mathcal{L}_{text} + \lambda_{img}\mathcal{L}_{img},
\end{equation}
where the weighting coefficients $\lambda_{text}$ and $\lambda_{img}$ of the generated text and image generation loss are set to 1 in our experiments.

\subsubsection{Controllable Advertisement Generation}
Vanilla autoregressive models fail to generate product-consistent advertising images and texts due to the lack of effective controls. To address this, we equip Uni-AdGen with separate control approaches designed for images and texts. For image generation, we design a foreground perception module that processes the product's transparent image, as illustrated in Fig.~\ref{fig:pipeline}(a). The transparent image is discretized into patches via patchify and projector layers, then encoded by the DINOv2-based foreground encoder \cite{dinov2} into visual embedding, which are further aligned to the autoregressive model's latent space using simple MLP-based projection layers, which we term control sequence layers.
Finally, the visual embedding is injected into the autoregressive decoder through element-wise addition at every 4 layers:
\begin{equation}
\left[\mathbf{H}_l\right]_t = \left[\text{DL}_l(\mathbf{H}_{l-1})\right]_t + \mathbb{I}_{l \bmod 4 = 0} \cdot \left[\mathbf{C}\right]_t,
\end{equation}
where, $\left[\mathbf{H}_l\right]_t$ denotes the input token at the $t$-th position of the $l$-th decoder layer, while $\text{DL}_l$ represents the $l$-th decoder layer. $\mathbf{C}$ is the aligned control signals from the foreground perception module. The indicator function $\mathbb{I}_{l \bmod 4 = 0}$ equals 1 when $l$ is divisible by 4 and 0 otherwise.

For advertising text generation, we employ instruction tuning to enhance the model's adherence to product information. Specifically, we design diverse instruction templates to automatically convert product descriptions and selling points into generation instructions, with one example formatted as: ``\textit{**Product Information** [Description]: \textit{\{\}}, [Selling Points]: \textit{\{\}}. **Task** Generate an advertising text and image based on [Description] and [Selling Points], where the text uses ONLY words from the provided [Selling Points].}". We then leverage an LLM to clean the training data by removing noisy samples where the ground-truth advertising texts cannot be inferred from the given selling points. This approach enables the model to learn a consistent generation strategy through instruction tuning.

\subsection{Personalized Ad Generation}
\label{sec:personalized_ad_gen}
\subsubsection{Product Similarity Sampling}
Building on the Uni-AdGen, we design a coarse-to-fine preference understanding module to use historical behaviors for personalized advertisement generation. As shown in Fig.~\ref{fig:pipeline}, the module first performs coarse-grained selection of the top-$N$ behaviors from a large pool of $M$ historical behaviors (where $M \gg N$) by product similarity for sample-level noise reduction.
Specifically, we measure the semantic similarity between historical products' texts and the target product description, then construct a candidate set by importance sampling from historical behaviors, with the sampling weights $p$:
\begin{equation}
   p_i = \frac{\max(s_i + \epsilon, 0)}{\sum_{j=1}^N \max(s_j + \epsilon, 0)},
\end{equation}
where $s_i$ represents the semantic similarity of the $i$-th product and $\epsilon$ is set to 1e-6 to prevent division by 0. Since products with similar semantics usually share advertising styles and textual patterns, the candidate set obtained by product similarity sampling can effectively reduce sample-level noise. Meanwhile, we empirically observe that certain historical products with lower semantic similarity still provide useful visual or textual references, such as the bookshelf and lipstick in Fig.~\ref{fig:problems}. With our sampling strategy, the candidate set probabilistically incorporates these items, thereby increasing reference diversity.

\begin{table*}
\caption{The quantitative results of general advertisement generation task. The best results are marked in \textbf{bold}, while the second-best results are \underline{underlined}. IR is ImageReward metric and PS is PickScore metric. The ``--" indicates experimental data that cannot be measured.}
\centering
\begin{adjustbox}{width=\linewidth}
\begin{tabular}
{l@{\hspace{0.5cm}}cccccccc}
\toprule
\multicolumn{1}{c}
{\multirow{3}{*}{\bf Method}} & \multicolumn{4}{c}{\bf Image Generation} & \multicolumn{3}{c}{\bf Text Generation} \\
\cmidrule(lr){2-5}\cmidrule(lr){6-8}
&
{\bf ASE $\uparrow$ } &
{\bf IR$\uparrow$} &
{\bf PS$\uparrow$} &
{\bf Human Eval$\uparrow$} &
{\bf m-BLEU$\uparrow$} &
{\bf m-ROUGE$\uparrow$} &
{\bf Human Eval$\uparrow$} & \\
\cmidrule{1-9}
\multicolumn{9}{c}{\cellcolor{gray!10}\textit{Image Generation Models}}\\
\cmidrule{1-9}
GPT-4o \cite{gpt4o} + ReliableAd \cite{reliablead} & 5.26 & -1.491 & 20.888 & \underline{93.80} & -- & -- & -- \\
Qwen2.5vl \cite{qwen2_5_vl} + ReliableAd \cite{reliablead} & 5.29 & -1.516 & 20.890 & \textbf{95.20} & -- & -- & -- \\
GPT-4o \cite{gpt4o} + PosterMaker \cite{postermaker} & \textbf{5.42} & -1.332 & 20.970 & 90.00 & -- & -- & -- \\
Qwen2.5vl \cite{qwen2_5_vl} + PosterMaker \cite{postermaker} & \underline{5.34} & -1.399 & 20.977 & 90.80 & -- & -- & -- \\
GPT-4o \cite{gpt4o} + Flux-Fill \cite{flux} & 5.21 & \underline{-1.281} & 20.926 & 88.00 & -- & -- & -- \\
Qwen2.5vl \cite{qwen2_5_vl} + Flux-Fill \cite{flux} & 5.14 & -1.311 & 20.918 & 86.80 & -- & -- & -- & \\
\textbf{Ours-image} & \textbf{5.42} & -1.351 & \textbf{21.081} & 92.40 & -- & -- & -- \\
\cmidrule{1-9}
\multicolumn{9}{c}{\cellcolor{gray!10}\textit{Text Generation Models}}\\
\cmidrule{1-9}
Qwen2.5 \cite{qwen3} &-- & -- & -- & -- & 0.530 & 0.651 & 97.00 \\
Qwen3 \cite{qwen2_5} & --& -- & -- & -- & \textbf{0.562} & {0.652} & \textbf{99.60} \\
DeepSeek-R1 \cite{deepseekr1} &  -- & -- & -- & -- & 0.533 & \underline{0.653} & 97.80 \\
\textbf{Ours-text} & -- & -- & -- & -- & \underline{0.559} & 0.650 & 97.80 \\
\cmidrule{1-9}
\multicolumn{9}{c}{\cellcolor{gray!10}\textit{Image and Text Joint Generation Models}}\\
\cmidrule{1-9}
\textbf{Ours} & 5.25 & \textbf{-1.244} & \underline{21.002} & 92.60 & 0.551 & \textbf{0.654} & \underline{98.20}\\
\bottomrule
\end{tabular}
\end{adjustbox}
\label{tab:non-personalized}
\end{table*}

\subsubsection{Multimodal Preference Extraction}
Although product similarity sampling can yield a highly relevant candidate set of historical behaviors, we still can not evaluate users' interests in a particular modality because of the modal-level noise. We introduce the multimodal preference extraction to alleviate the modal-level noise. As shown in Fig.~\ref{fig:pipeline}(b), the sampled $N$ historical image-text pairs are first encoded into token sequences using image and text encoders. These tokens are subsequently prepared for multimodal preference extraction, where two separate Transformer encoders \cite{transformer} serve as relevance extractor for visual and textual tokens. Each extractor measures token importance through attention mechanisms and generates token-level masks by computing cosine similarity between input and output embeddings, followed by a differentiable Gumbel-Softmax $\text{G}(\cdot)$ with top-K selection $\text{TK}(\cdot)$ to preserve relevant content while suppressing modal-level noise:
\begin{equation}
    \mathbf{e}^i = \text{TK}\left(\text{G}\left(\mathbf{e}_{in}^i \cdot \mathbf{e}_{out}^i\right)\right) \cdot \mathbf{e}_{in}^i\,\,\,\,i\in\{v, t\},
\end{equation}
where $\mathbf{e}_{in}^i$ and $\mathbf{e}_{out}^i$ denote the input and output tokens of the relevance extractor for $i$, with $v$ and $t$ representing the visual and textual modalities respectively. The resulting tokens $\mathbf{e}^i$ are further processed to obtain the fused image tokens $\mathbf{e}_{fused}^{v}$ and fused text tokens $\mathbf{e}_{fused}^{t}$:
\begin{equation}
     [\mathbf{e}_{fused}^v, \mathbf{e}_{fused}^t] = PF\left((f(\mathbf{e}^v) + \mathbf{e}^v)\oplus(f(\mathbf{e}^t) + \mathbf{e}^t)\right),
\end{equation}
where $f(\cdot)$ represents the feedforward layers (FFN) with residual connections, and $PF(\cdot)$ denotes another Transformer encoder to fuse the multimodal tokens after concatenation operation $\oplus$. Finally, we insert the fused tokens into the extended instruction to drive personalization by appending a style constraint ``\textit{Generation content following the text style $\texttt{<text\_ph>}$ and the image style $\texttt{<image\_ph>}$}" at the end, where the placeholders $\texttt{<image\_ph>}$ and $\texttt{<text\_ph>}$ specify the insertion positions.

\section{Experiment}
\label{sec:experiment}
\subsection{Implementation Details}
Our method builds upon the Janus-Pro 7B \cite{Janus-Pro}. 
We train our model with a batch size of 4 using the AdamW optimizer (lr=5e-5). The base model is fine-tuned via LoRA \cite{lora} (rank=12, factor=32), while the foreground perception module and multimodal preference extraction are fully fine-tuned.
We sample $N=10$ historical behaviors in the product similarity sampling and keep the top 40\% tokens in the multimodal preference extraction. All experiments are run on a single NVIDIA B200 GPU, with the transparent and historical images resized to $384\times384$ and $128\times128$. 

\subsection{General Ad Generation Performance}
\label{sec:general_ad_gen_exp}

\begin{figure*}[ht]
    \centering
    \includegraphics[width=\linewidth]{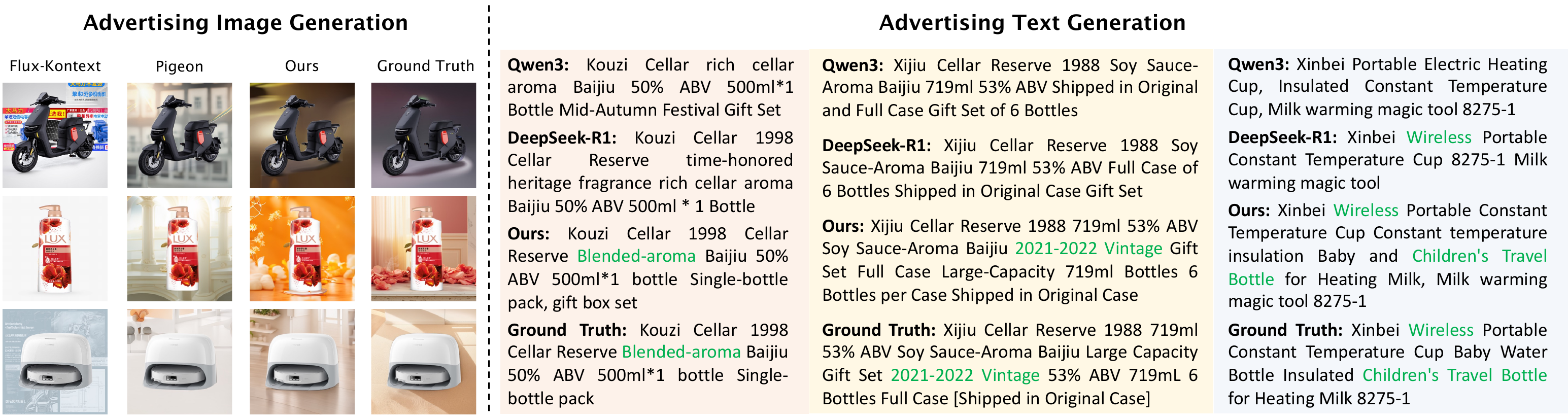}
    \caption{The qualitative results of personalized advertisement generation task. Our method generates images with colors and scenes closer to actual clicks, while the produced texts cover more selling points from the ground truth (highlighted in \textcolor[rgb]{0, 0.69, 0.31}{green}).}
    \label{fig:personalized_ad_gen}
\end{figure*}

\noindent\textbf{Metrics.} For advertising image generation, we employ PickScore \cite{pickscore}, ImageReward \cite{imagereward}, and ASE \cite{ase} to quantify generation quality from the aesthetic perspective. For advertising text generation, we adopt m-BLEU \cite{bleu} and m-ROUGE \cite{rouge} to evaluate the similarity under a multi-candidate setting. The prefix ``m-" denotes that each generated content is compared against multiple candidate texts, with final scores averaged over 1- to 4-gram matches. We also conduct human evaluation to assess the available rate of generated content in e-commerce.  For images, we examine product structural consistency and layout rationality following \cite{reliablead}. For texts, we focus on factual accuracy and linguistic fluency. More details about human evaluation are provided in the appendix. All metrics are computed on 500 products.


\noindent\textbf{Baselines.} 
We construct advertising image generation baselines following the pipeline mentioned in Sec.~\ref{sec:intro}. We adopt Qwen2.5-VL \cite{qwen2_5_vl} and GPT-4o \cite{gpt4o} to create background prompts from product transparent images. These prompts serve as input to two e-commerce-oriented methods (ReliableAd \cite{reliablead} and PosterMaker \cite{postermaker}), along with the SOTA model Flux-Fill \cite{flux}, to obtain final advertising images. For text generation, we employ the same experimental setup and select a set of leading LLMs as baselines, including Qwen2.5 \cite{qwen2_5}, Qwen3 \cite{qwen3}, and DeepSeek-R1 \cite{deepseekr1}.

\noindent\textbf{Performance.} 
We present quantitative results in Table \ref{tab:non-personalized}. As shown in Table~\ref{tab:non-personalized}, our method achieves the best performance in ImageReward and ranks second in both PickScore and human evaluation, demonstrating its superior performance in aesthetic and high available rate. While ReliableAd leads in human evaluation, it lags significantly in aesthetic metrics. Conversely, PosterMaker and Flux-Fill generate visually appealing images but suffer from noticeable usability limitations. Thanks to effective control approaches, our method successfully achieves an optimal balance between visual content and practical utility. In advertising text generation, our method achieves second-best performance in human evaluation and shows competitive performance comparable to state-of-the-art baselines on metrics including m-BLEU and m-ROUGE. 

Additionally, we train single-modal variants of our model (Ours-image, Ours-text) to evaluate their performance on individual modality generation tasks. Results show that our full model performs comparably to these variants, confirming the effectiveness of multi-modal training.

\subsection{Personalized Ad Generation Performance}

\noindent\textbf{Metrics.} 
We evaluate on 500 users with historical behaviors, using PBS (Sec.~\ref{sec:dataset}) for image similarity and BLEU/ROUGE for text similarity against actual clicks.

\noindent\textbf{Baselines.} 
We select Flux-Kontext \cite{flux_kontext} and Pigeon \cite{pigeon} for personalized image generation, as the image generation baselines in Sec.~\ref{sec:general_ad_gen_exp} cannot perceive user history. For text generation, we adopt Qwen3 \cite{qwen3} and DeepSeek-R1 \cite{deepseekr1}.
When using Flux-Kontext, we arrange historical images into a grid layout and feed it into the model alongside the target product's transparent image, enabling historical content to guide the generation. 
Since Pigeon does not natively support controllable generation, we integrate our foreground perception module to equip it with product-consistent generation capability. For Qwen3 and DeepSeek-R1, we insert historical product texts directly into the instruction template.

\noindent\textbf{Performance.} In Table~\ref{tab:personalized}, our method achieves superior performance in both image and text personalized generation, significantly outperforming baselines.  The visualized results in Fig.~\ref{fig:personalized_ad_gen} show that Flux-Kontext fails to understand user preferences and remains susceptible to sample-level noise, resulting in significant deviation from ground truth, such as the irrelevant items in the motorcycle image. Limited by its single-modal design, Pigeon achieves suboptimal performance due to inaccurate preference modeling. Similarly, texts generated by Qwen3 and DeepSeek-R1 exhibit lower coverage of selling points present in real clicks, such as missing ``Blended-aroma". In contrast, our coarse-to-fine preference understanding module jointly models user interests from image-text history, enabling Uni-AdGen to generate advertisement that better align with actual clicks. It is worth noting that even a small increase in advertising CTR can translate into significant revenue, highlighting our method’s practical value. 

We conduct ablation studies by progressively enhancing Uni-AdGen without history (baseline), with random historical data (w/ history), product similarity sampling (w/ PSS), and finally multimodal preference extraction (Ours).
Results in Table~\ref{tab:personalized} reveal that the baseline exhibits significant gaps from users' actual clicks. Adding random historical behaviors (w/ history) obtains marked improvements across all metrics, confirming the value of historical data for personalization. The PSS-enhanced model brings further gains by reducing sample-level noise through relevant behavior selection. Our full model, equipped with multimodal preference extraction, best aligns with user preferences by capturing fine-grained interests, achieving optimal performance. Additional results are provided in the appendix.
\begin{table}[tp]
\centering
\caption{Ablation studies for personalized generation. ``Baseline" indicates Uni-AdGen without historical data; ``w/ history" adds historical behaviors; ``w/ PSS" further includes product similarity sampling; and ``Ours" is the complete model.}
\setlength{\abovecaptionskip}{4pt}
\resizebox{1\linewidth}{!}{
\setlength{\tabcolsep}{5 mm}
\begin{tabular}{lccc}  
\toprule
{\bf Method} &
{\bf PBS $\uparrow$} &
{\bf BLEU$\uparrow$} &
{\bf ROUGE$\uparrow$}\\
\cmidrule{1-4}
\multicolumn{4}{c}{\cellcolor{gray!10}\textit{Image Generation Models}}\\
\cmidrule{1-4}
Pigeon \cite{pigeon} & 0.624 & -- & -- \\
Flux-Kontext \cite{flux_kontext} & 0.514 & -- & -- \\
\cmidrule{1-4}
\multicolumn{4}{c}{\cellcolor{gray!10}\textit{Text Generation Models}}\\
\cmidrule{1-4}
Qwen3 \cite{qwen3} & -- & 0.345 & 0.580\\
DeepSeek-R1 \cite{deepseekr1} & -- & 0.373 & 0.622 \\
\cmidrule{1-4}
\multicolumn{4}{c}{\cellcolor{gray!10}\textit{Image and Text Joint Generation Models}}\\
\cmidrule{1-4}
Baseline & 0.617 & 0.225 & 0.525 \\
w/ history & 0.606 & 0.427 & 0.650 \\
w/ PSS & 0.622 & 0.430 & 0.652 \\
\textbf{Ours} & \textbf{0.634} & \textbf{0.435} & \textbf{0.662} \\
\bottomrule
\end{tabular}}
\label{tab:personalized}
\end{table}
\section{Conclusion}
\label{sec:conclusion}
This paper presents Uni-AdGen, a unified autoregressive framework that jointly produces realistic image-text advertisement. To enable personalization, we design a coarse-to-fine preference understanding module to model accurate user interests from noisy behavioral data. We further contribute PAd1M, the first large-scale personalized advertising image-text dataset, to support model training and evaluation. Extensive experiments validate the effectiveness of our method in both general and personalized settings.
\section*{Acknowledgements}
This work was partially supported by the Guangdong S\&T Programme (No. 2025B0101130003), the Guangdong Basic and Applied Basic Research Foundation (2022B1515020103, 2023B1515120087), and the Science and Technology Planning Project of Key Laboratory of Advanced IntelliSense Technology, Guangdong Science and Technology Department (2023B1212060024). Thanks to Yanyin Chen for her insightful discussions and support for this paper.


    \small
    \bibliographystyle{ieeenat_fullname}
    \bibliography{main}

\clearpage
\setcounter{page}{1}
\maketitlesupplementary

This appendix details the design for human evaluation metrics in Sec.~\ref{supp:human_eval}. Section~\ref{supp:visualized} provides comprehensive visual comparisons with baseline methods under both personalized and non-personalized settings. To justify our model configuration, Section~\ref{supp:ablation} systematically analyzes how historical length and sampling strategies affect performance. In Sec.~\ref{supp:instruction_template}, we present the instruction tuning templates used in this study. Finally, Section~\ref{supp:limitation} and~\ref{supp:social_impact} discuss technical limitations and social impact respectively, ensuring our research advances technology while maintaining ethical and social responsibility.

\section{Details of Human Evaluation Metric}
\label{supp:human_eval}
To evaluate whether the generated contents (advertising images and text) comply with e-commerce standards, we established strict human evaluation criteria shown in Sec.~\ref{sec:supp_image_criteria} and Sec.~\ref{sec:supp_text_criteria}.
\subsection{Advertising Image Generation}
\label{sec:supp_image_criteria}
 The human evaluation of generated advertising images is aligned with \cite{reliablead} and assesses four key aspects: product size, product appearance, spatial positioning, and visual perception. Any violation of these criteria results in image rejection. The metric reports the percentage of qualified images among all generated images.

\begin{figure}[h]
    \centering
    \includegraphics[width=\linewidth]{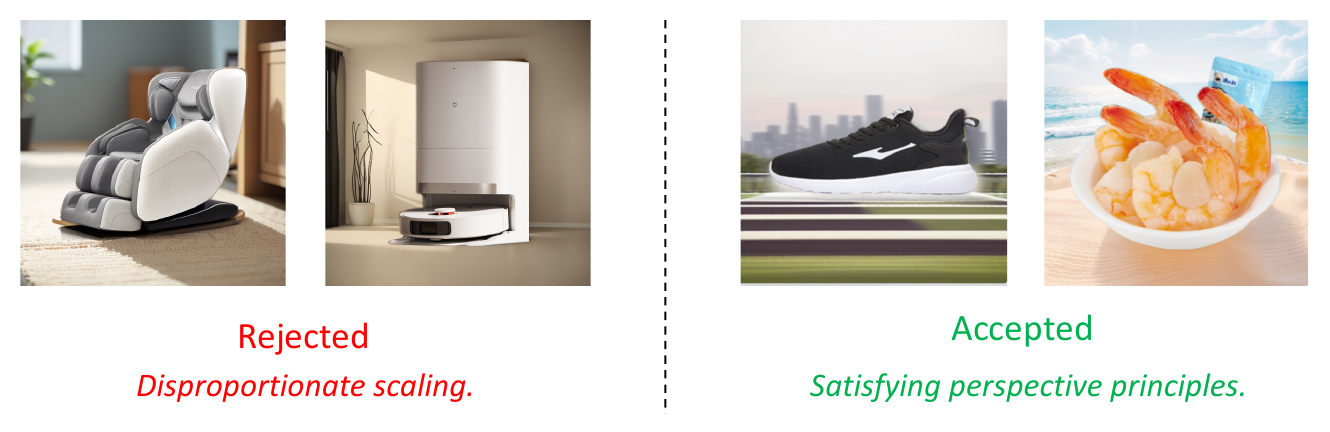}
    \caption{The examples of size mismatch.}
    \label{fig:product_size}
\end{figure}

\noindent\textbf{Size mismatch.}
As shown in Fig.~\ref{fig:product_size}, generated advertising images must maintain a reasonable proportion between products and indoor reference objects (e.g., ceilings, tables, baseboards). Disproportionate scaling leads to rejection. For outdoor scenes, images satisfying perspective principles (e.g., nearby objects appearing larger) are considered acceptable.

\begin{figure}[h]
    \centering
    \includegraphics[width=\linewidth]{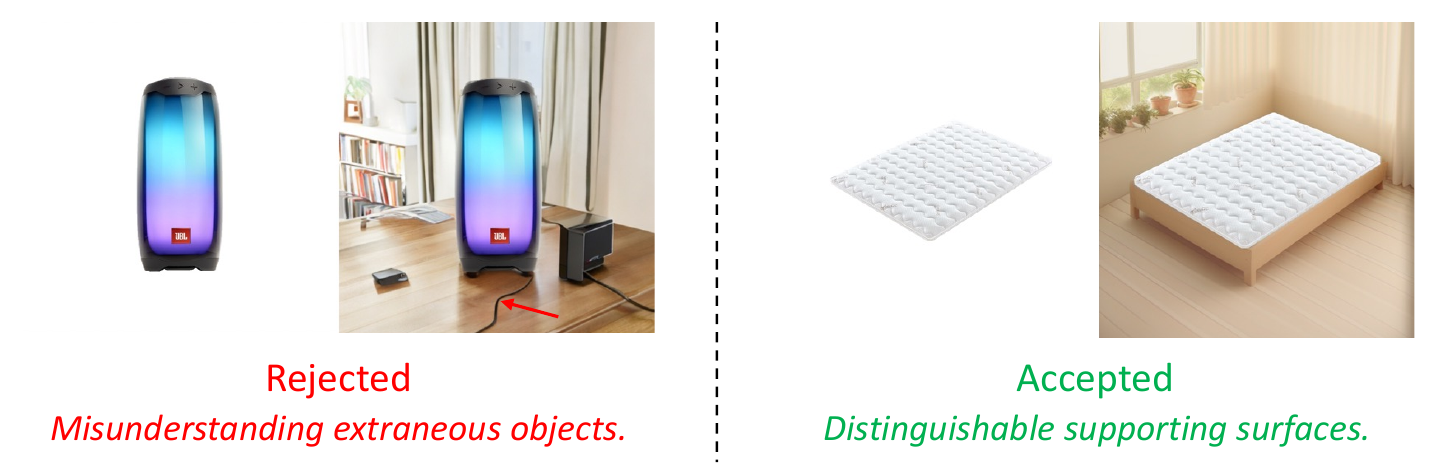}
    \caption{The examples of shape hallucination.}
    \label{fig:product_appearance}
\end{figure}

\noindent\textbf{Shape hallucination.}
Images are rejected if extraneous objects attached to product edges could cause user misunderstanding. For example, the wireless speaker appears as a wired speaker in Fig.~\ref{fig:product_appearance}. However, clearly distinguishable supporting surfaces (e.g. products on stands) are acceptable. For hollow areas: minor artifacts are tolerated, but significant mismatches between hollow areas and background colors result in rejection.

\begin{figure}[h]
    \centering
    \includegraphics[width=\linewidth]{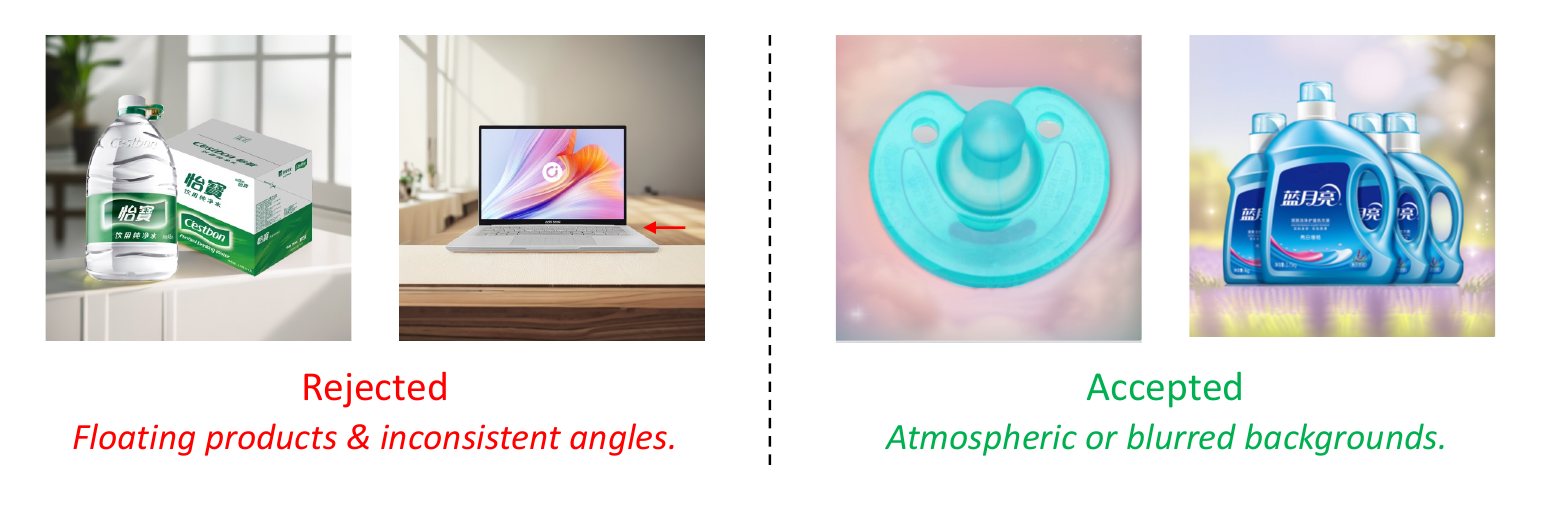}
    \caption{The examples of space mismatch.}
    \label{fig:spatial_positioning}
\end{figure}

\noindent\textbf{Space mismatch.}
As illustrated in Fig.~\ref{fig:spatial_positioning}, images are rejected if products appear spatially disconnected from backgrounds, creating unrealistic impressions, including floating products or inconsistent angles between products and backgrounds. Images with atmospheric or blurred backgrounds are exempt from spatial consistency requirements.

\begin{figure}[h]
    \centering
    \includegraphics[width=\linewidth]{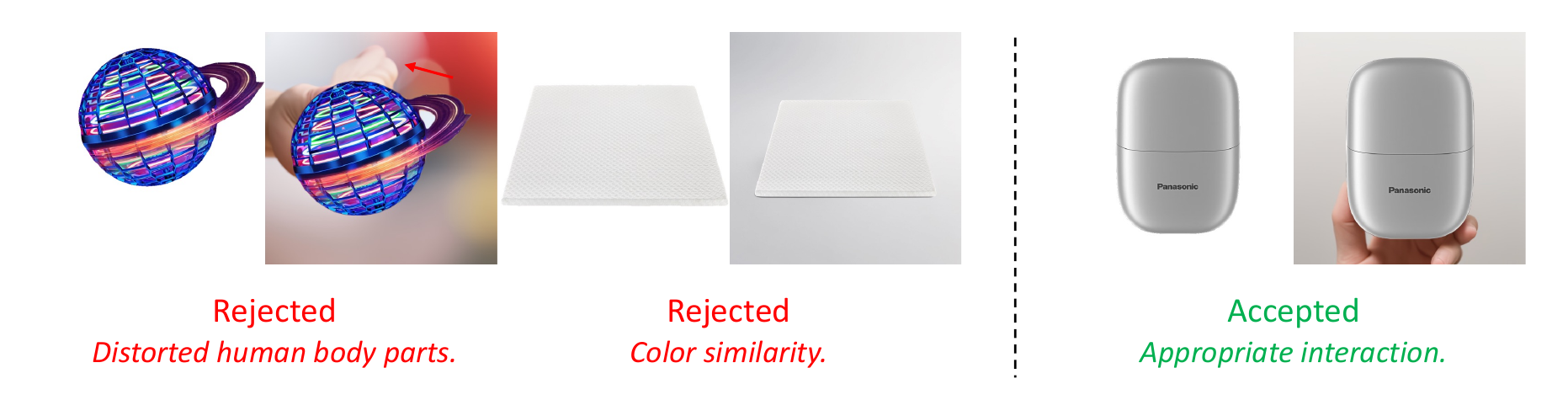}
    \caption{The examples of visual perception.}
    \label{fig:visual_perception}
\end{figure}

\noindent\textbf{Visual perception.}
This criterion evaluates product distinctiveness and potential discomfort. Images containing distorted human body parts (e.g., faces, hands) that cause discomfort or inappropriate associations are rejected, unless bodies naturally interact with products shown in Fig.~\ref{fig:visual_perception}. Backgrounds that obscure products due to color similarity or excessive complexity also lead to rejection.

\begin{figure*}[t]
    \centering
    \includegraphics[width=\linewidth]{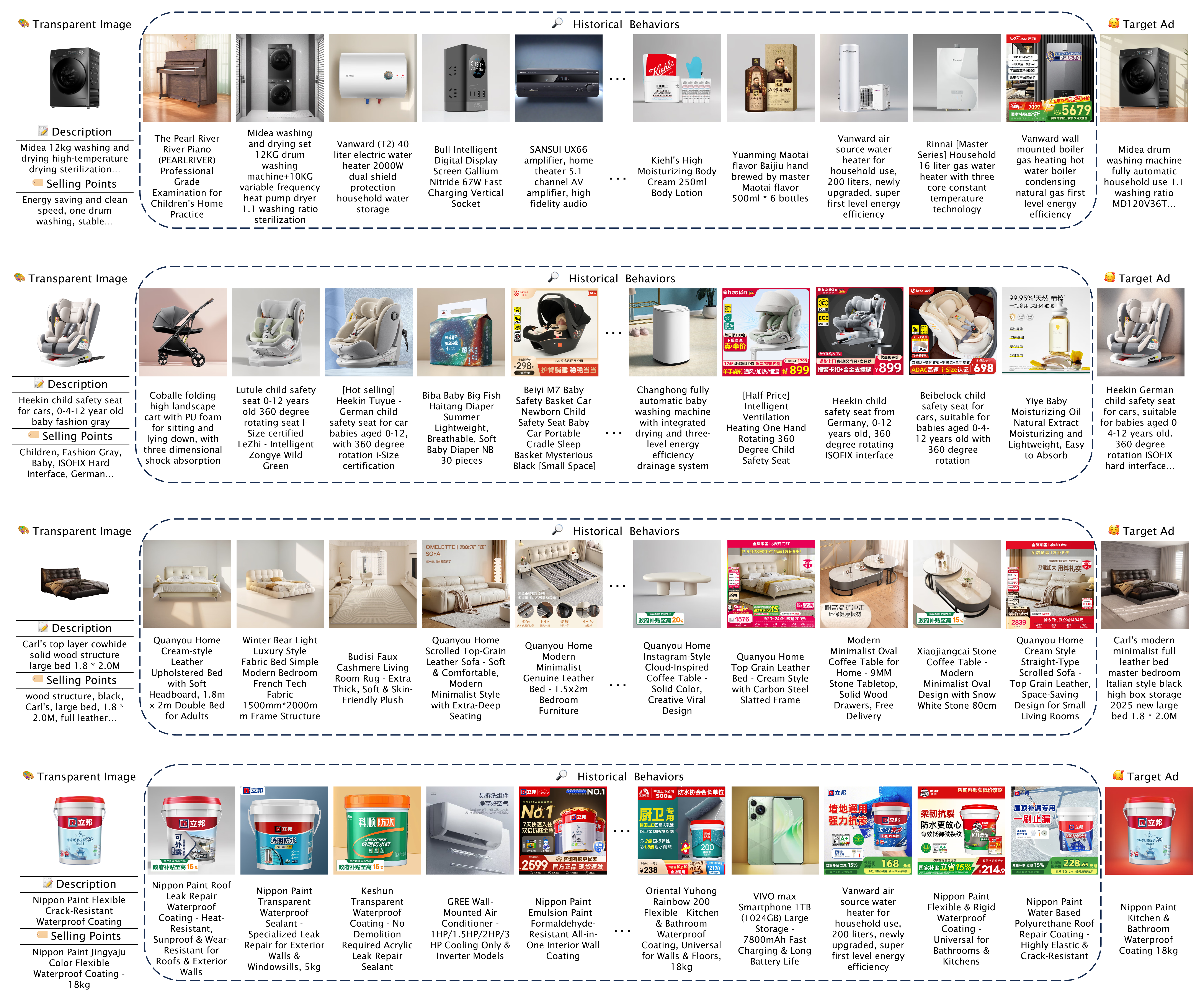}
    \caption{More visualization examples in the PAd1M dataset.}
    \label{fig:supp_dataset}
\end{figure*}

\subsection{Advertising Text Generation}
\label{sec:supp_text_criteria}
\begin{figure}
    \centering
    \includegraphics[width=\linewidth]{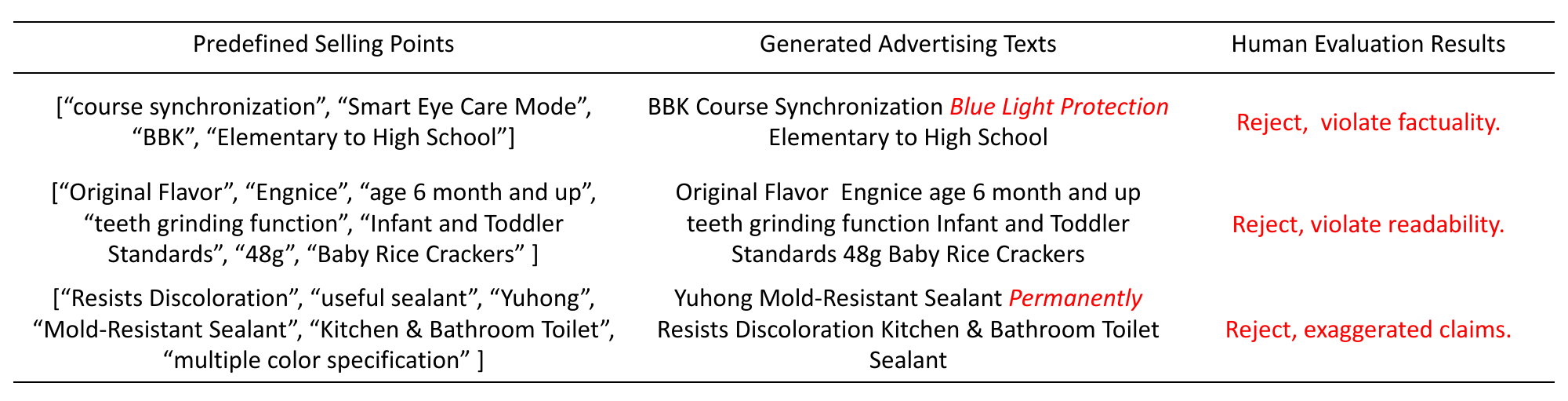}
    \caption{The examples of human evaluation for advertising texts.}
    \label{fig:human_evaluation_texts}
\end{figure}

Generated advertising texts are evaluated based on factuality and readability as shown in Fig.~\ref{fig:human_evaluation_texts}. Violation of either criterion results in rejection. The metric reports the percentage of qualified texts.

\noindent\textbf{Factuality.}
Texts are rejected if they contain selling points absent from the predefined list. Inclusion of sensitive content (e.g., pornography, violence) or exaggerated/ false claims also leads to rejection.

\noindent\textbf{Readability.}
Texts fail if they contain incoherent sentences, direct copying of selling points, or severe repetition of selling points.

\subsection{Annotator Details}
To ensure the reliability of our annotated data, we implemented a rigorous three-stage manual review process. The annotations were performed by three independent teams of domain experts, each possessing over five years of relevant industry and regulatory experience. In the first stage, annotators performed initial labeling based on predefined guidelines. The second stage involved a verification step, where a separate group of reviewers checked the labels for consistency and correctness. Finally, a random recheck was conducted on a subset of the data to identify any remaining discrepancies. This multi-tiered quality control guarantees an annotation accuracy of at least 98$\%$.

\begin{figure*}[t]
    \centering
    \includegraphics[width=0.95\linewidth]{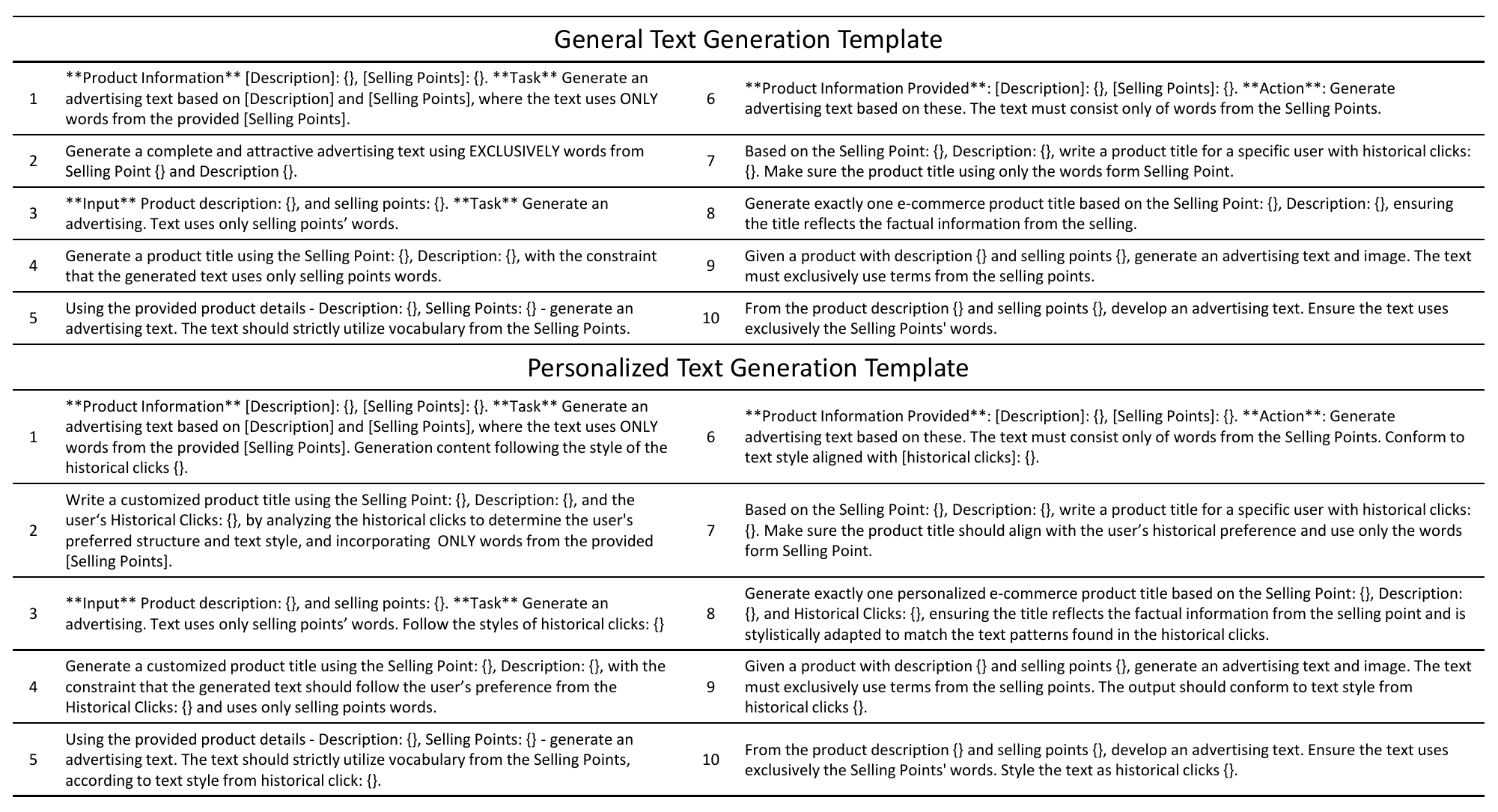}
    \caption{The template pools for advertising text generation task.}
    \label{fig:text_prompts}
\end{figure*}
\begin{figure*}
    \centering
    \includegraphics[width=0.95\linewidth]{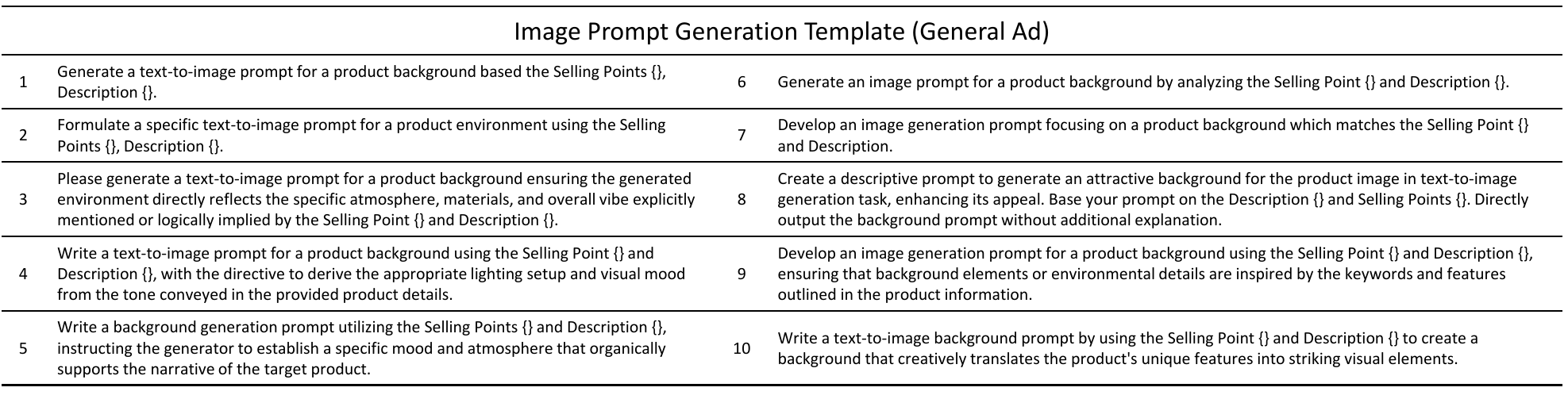}
    \caption{The template pools for image prompt generation.}
    \label{fig:image_prompts}
\end{figure*}

\section{Baseline Implementation Details}
To ensure fair comparison, prompts for the text generation task are randomly sampled from a diverse template pool shown in Fig.~\ref{fig:text_prompts}, while image prompts are generated by QwenVL/GPT‑4o using various templates in Fig.~\ref{fig:image_prompts} to ensure stylistic variety.

\section{Additional Visualizations}
\label{supp:visualized}
In this section, we first present additional examples from our Personalized Advertising image-text dataset (PAd1M) to illustrate its composition. We then provide more qualitative results across methods for both general and personalized advertisement generation, demonstrating the advantages of our method in jointly generating advertising images and texts.

\subsection{Personalized Advertising Image-text Dataset}
Figure~\ref{fig:supp_dataset} presents additional examples from the PAd1M dataset, with each row showing a complete user record. Every sample includes the target product's transparent image, textual description, and selling points, together with its corresponding ground-truth advertising image and title. Additionally, we preserve the complete sequence of each user's historical click behaviors, where every historically clicked product is accompanied by its original advertising image and text.

\subsection{General Ad Generation}
\begin{figure*}
    \centering
    \includegraphics[width=0.95\linewidth]{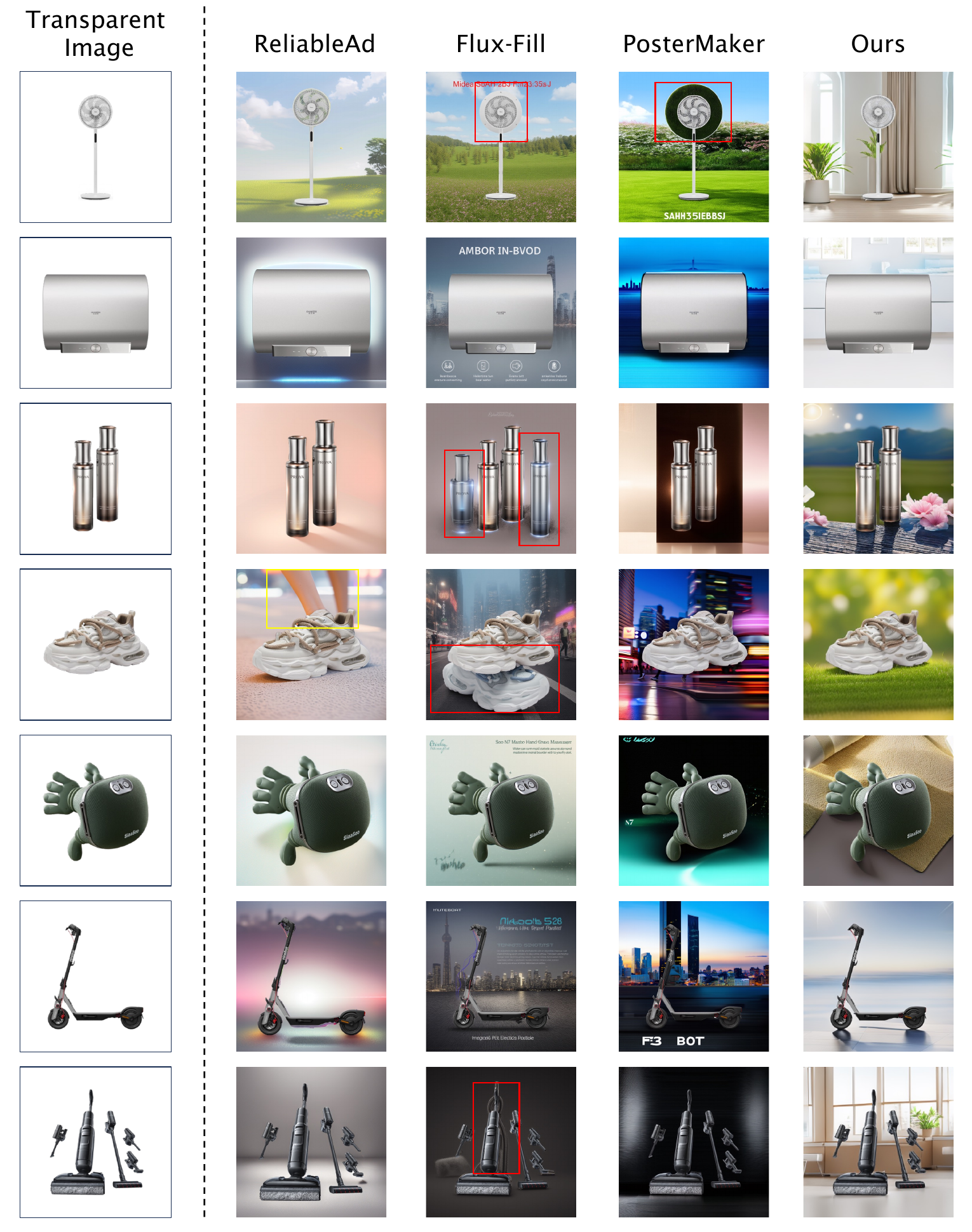}
    \caption{The generated advertising images of different methods on general advertisement generation tasks. The \textcolor{red}{red} boxes mark misleading extraneous objects, and the \textcolor{yellow}{yellow} boxes mark uncomfortable generated results (such as incomplete body parts).}
    \label{fig:supp_non_personalized_wbaselines}
\end{figure*}

Figure~\ref{fig:supp_non_personalized_wbaselines} presents a visual comparison between our method and current state-of-the-art approaches. It can be observed that Flux-Fill \cite{flux} and PosterMaker \cite{postermaker} generate advertising images with richer color and higher visual realism in background design, which explains their advantages in aesthetic metrics. However, these images often contain misleading extraneous objects or unexplainable characters on the main products, compromising the practical usability of the generated results. In contrast, ReliableAd \cite{reliablead} achieves a higher available rate by employing structurally simple backgrounds with minimal color composition at the cost of limited visual appeal, resulting in inferior aesthetic scores. With the effective control approaches, our method maintains a high available rate while generating more realistic background content, achieving an optimal balance between aesthetic quality and practical value. 

\begin{figure*}
    \centering
    \includegraphics[width=\linewidth]{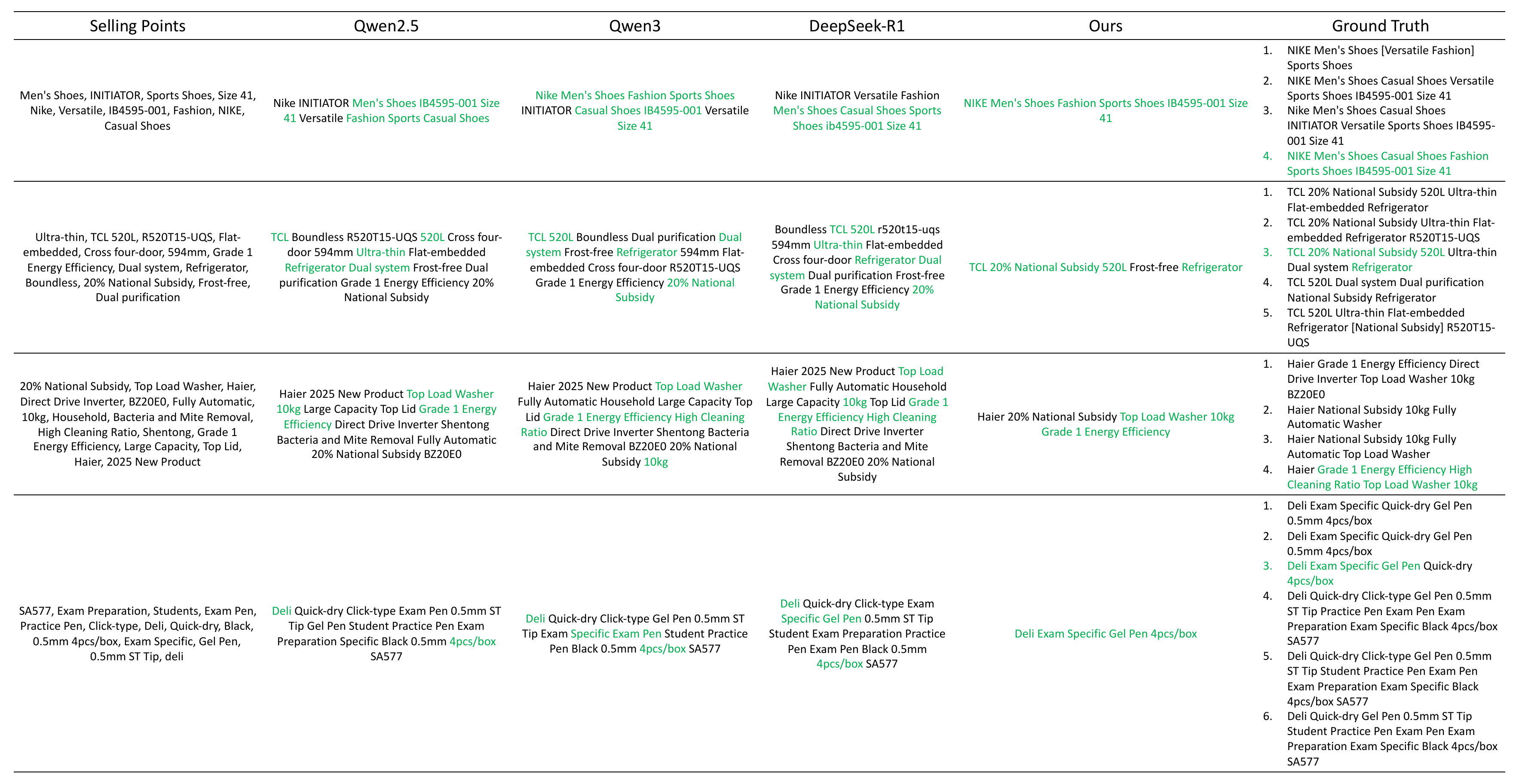}
    \caption{The generated advertising texts of different methods on general advertisement generation tasks. Some of the covered selling points are marked in \textcolor{green}{green}.}
    \label{fig:supp_non_personalized_wbaselines_text}
\end{figure*}

\begin{figure*}
    \centering
    \includegraphics[width=0.9\linewidth]{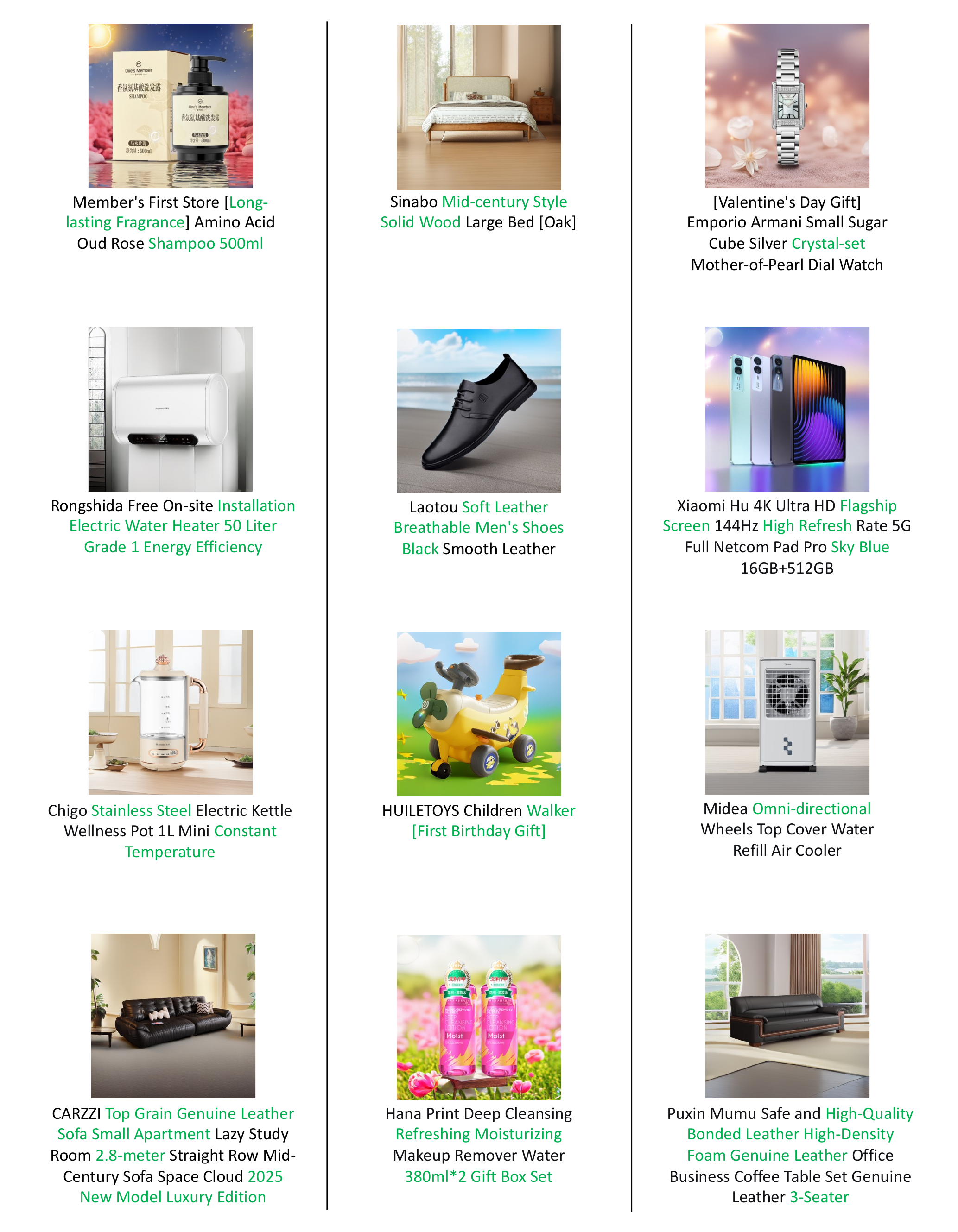}
    \caption{The generated advertising images and texts of our method on general advertisement generation tasks. Some of the covered selling points are marked in \textcolor{green}{green}.}
    \label{fig:supp_non_personalized_ours}
\end{figure*}

We present a comparison of advertising text generation results in Fig.~\ref{fig:supp_non_personalized_wbaselines_text}. It reveals that existing methods \cite{deepseekr1, qwen2_5, qwen3} tend to extensively copy selling points, which contributes to their strong performance on metrics like BLEU and ROUGE, yet leads to lower human evaluation scores due to poor semantic organization. In contrast, our approach employs instruction tuning and semantic constraints to ensure generated texts not only faithfully preserve selling points but also achieve natural fluency and professional organization comparable to ground-truth advertising texts. Additional examples in Fig.~\ref{fig:supp_non_personalized_ours} further validate our method's effectiveness.

\subsection{Personalized Ad Generation}
\begin{figure*}
    \centering
    \includegraphics[width=0.95\linewidth]{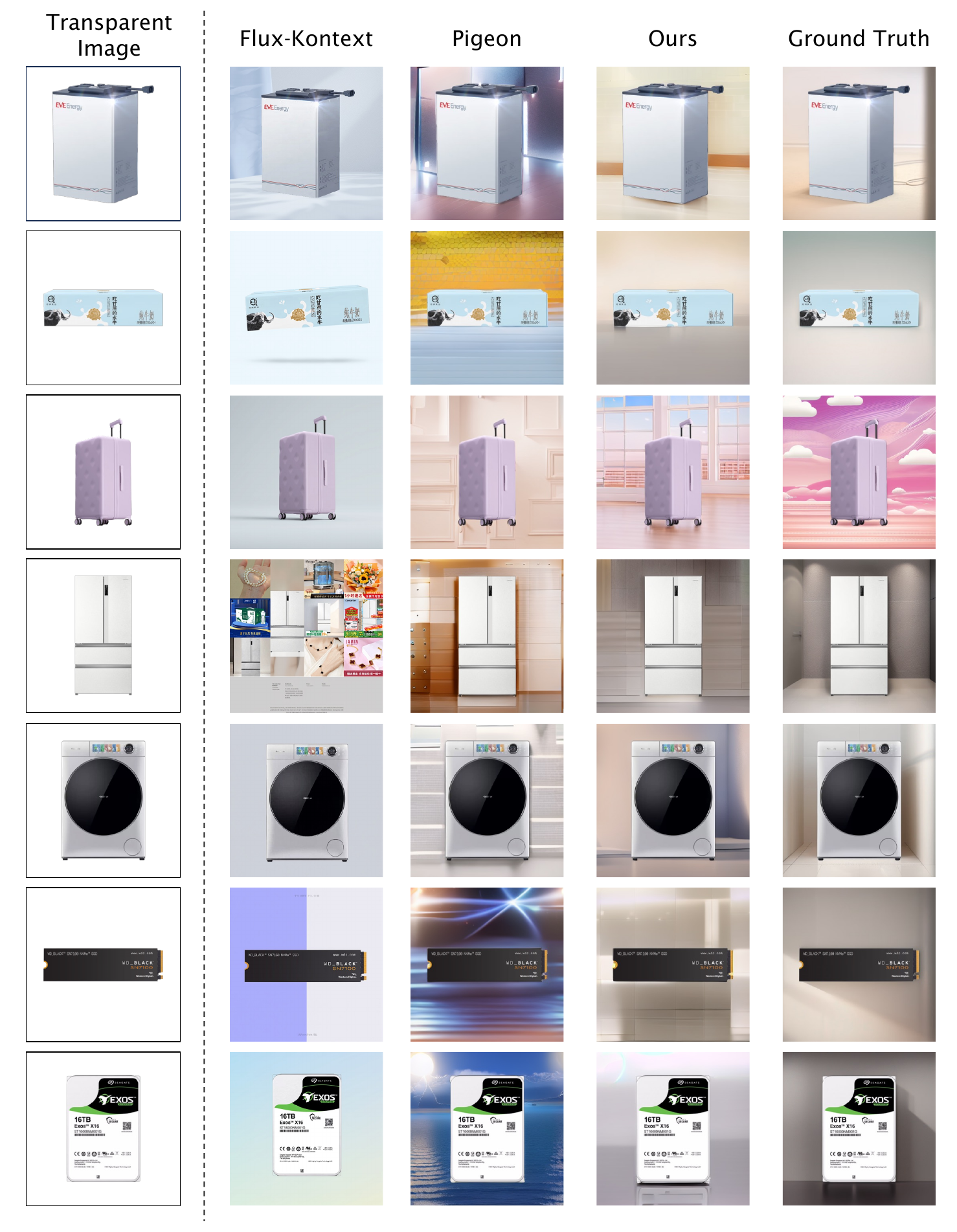}
    \caption{The generated advertising images of different methods on personalized advertisement generation tasks.}
    \label{fig:supp_personalized_wbaselines}
\end{figure*}

Figure \ref{fig:supp_personalized_wbaselines} presents comparative results of different methods on personalized image generation. It can be observed that Flux-Kontext \cite{flux_kontext} struggles to maintain proper product proportions in generated images (e.g. row 2 in Fig.\ref{fig:supp_personalized_wbaselines}) and tends to directly replicate historical content rather than extracting user preferences, resulting in significant deviations from actual clicks. While Pigeon \cite{pigeon} shows improved generated images through its preference extraction module, the single-modal architecture remains susceptible to sample-level noise. For instance, in the refrigerator case (row 4 in Fig.~\ref{fig:supp_personalized_wbaselines}), although both Pigeon and ground-truth feature indoor backgrounds, noticeable discrepancies persist in color selection. In contrast, our coarse-to-fine preference understanding module effectively reduces the negative effect from noise in historical behaviors, enabling generated images to better align with user preferences in both scene and color style.

\begin{figure*}
    \centering
    \includegraphics[width=\linewidth]{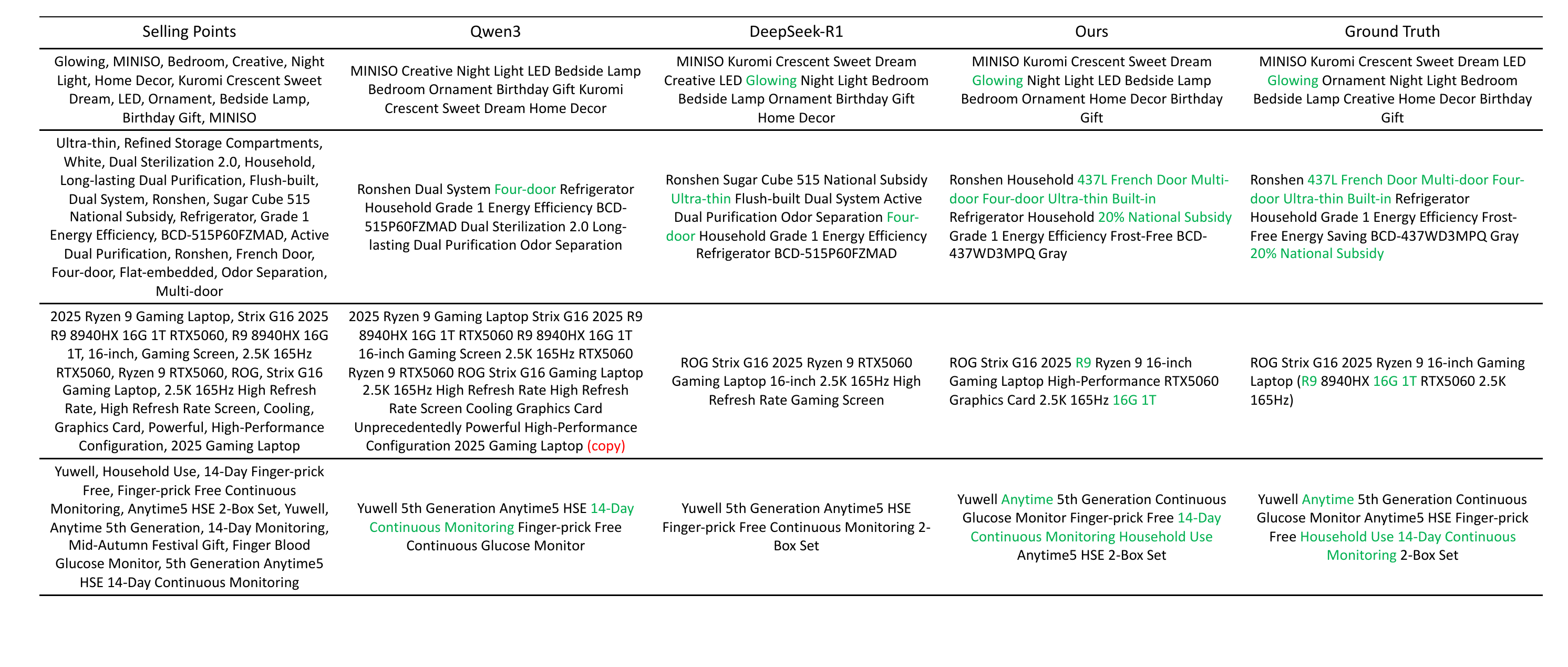}
    \caption{The generated advertising texts of different methods on personalized advertisement generation tasks. Some of the covered selling points are marked in \textcolor{green}{green}.}
    \label{fig:supp_personalized_wbaselines_text}
\end{figure*}

\begin{figure*}
    \centering
    \includegraphics[width=0.9\linewidth]{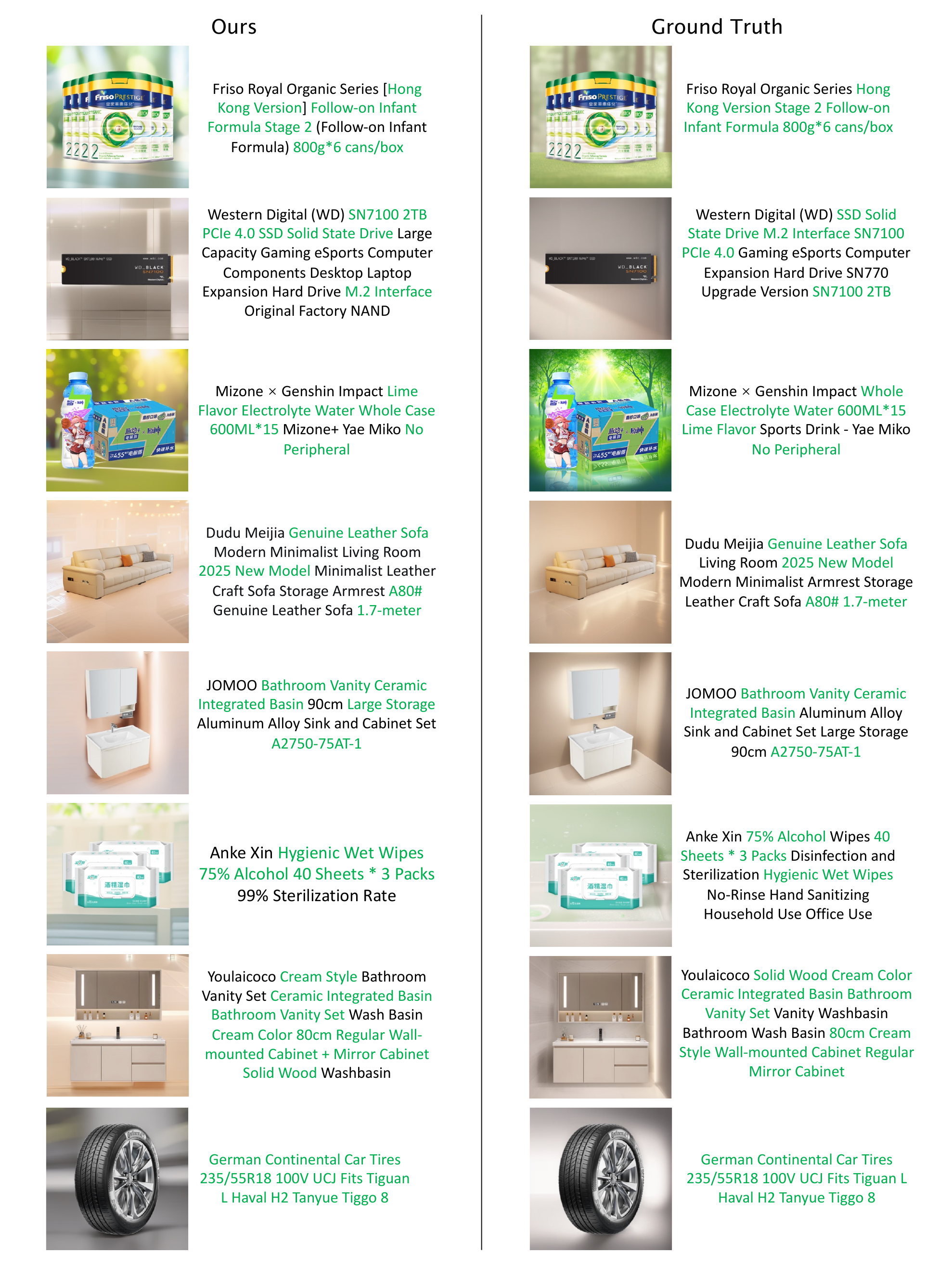}
    \caption{The generated advertising images and texts of our method on personalized advertisement generation tasks. Some of the covered selling points are marked in \textcolor{green}{green}.}
    \label{fig:supp_personalized_ours}
\end{figure*}

We compare text generation outputs across different methods in Fig.~\ref{fig:supp_personalized_wbaselines_text}. It can be observed that Qwen3 tends to directly copy selling points while DeepSeek-R1 shows moderate improvement in this aspect. Our method shows not only more comprehensive coverage of selling points but also closer sentence structures and expressive styles to real clicks. These results validate the advantage of our approach in modeling user preferences. More generated texts are shown in Fig.~\ref{fig:supp_personalized_ours}.

\begin{figure*}[ht]
    \centering
    \includegraphics[width=\linewidth]{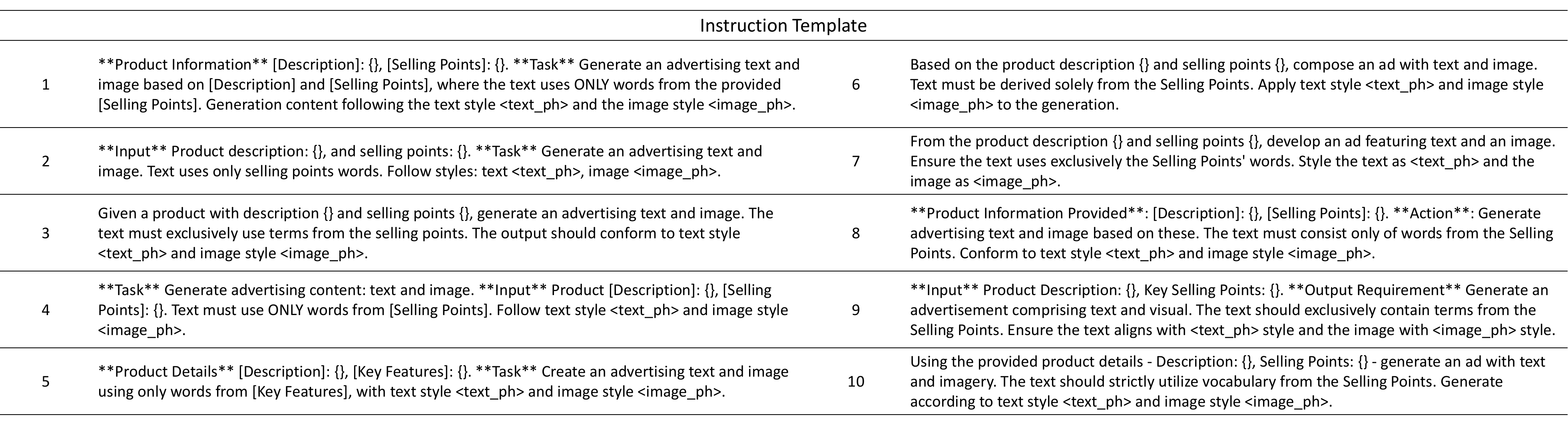}
    \caption{The instruction template used for instruction tuning.}
    \label{fig:instruction_template}
\end{figure*}

\section{Additional Results}
\label{supp:ablation}
\noindent\textbf{History length.} We examine how historical behavior sequence length affects model performance, and the results are shown in Table \ref{tab:history_length}. Without historical data, the model generates content that significantly diverges from ground-truth clicks and results in poor performance across all three evaluation metrics due to the absence of user preference guidance. As the history length increases, the model progressively extracts more complete user preferences, leading to steady improvements on metrics, where the best performance for image and text generation is reached at history lengths of 5 and 10, respectively. However, further extending the sequence introduces more noise than useful information, causing performance degradation. Based on these findings, we select a history length of 10 as the final configuration for balanced performance in image and text generation. 
\begin{table}[t]
\setlength{\abovecaptionskip}{4pt}
\caption{The result of different history length setting.}
\resizebox{1\linewidth}{!}{
\setlength{\tabcolsep}{3 mm}
\begin{tabular}{c|ccc}
\toprule
History Length & PBS $\uparrow$  & BLEU $\uparrow$ & ROUGE $\uparrow$\\ \midrule
0 & 0.617 & 0.225 & 0.525 \\
3 & 0.631 & 0.375 & 0.615 \\
5 & \textbf{0.639} & 0.400 & 0.637 \\ 
8 & 0.623 & 0.414 & 0.650 \\
10 & 0.634 & \textbf{0.435} & \textbf{0.662} \\
12 & 0.624 & 0.417 & 0.644 \\
\bottomrule
\end{tabular}
}
\label{tab:history_length}
\end{table}

\begin{table}[t]
\setlength{\abovecaptionskip}{4pt}
\caption{The results of different sampling strategies.}
\resizebox{1\linewidth}{!}{
\setlength{\tabcolsep}{3 mm}
\begin{tabular}{c|ccc}
\toprule
Setting & PBS $\uparrow$  & BLEU $\uparrow$ & ROUGE $\uparrow$\\ \midrule
Random Sampling & 0.610 & 0.431 & 0.656 \\
Most-similar Sampling & 0.618 & \textbf{0.439} & \textbf{0.663} \\
Product Similarity Sampling & \textbf{0.634} & 0.435 & 0.662 \\ 
\bottomrule
\end{tabular}
}
\label{tab:selection_strategy}
\end{table}

\noindent\textbf{Sampling strategy.} In Table \ref{tab:selection_strategy}, we compare the impact of different historical behavior sampling strategies on model performance. Specifically, we test three sampling methods on our full model: ``Random Sampling" chooses N historical behaviors randomly, ``Most-Similar Sampling" directly picks the top-N behaviors with the highest product similarity, and ``Product Similarity Sampling" refers to our proposed approach. Results show that random sampling obtains inferior performance due to noise in historical behaviors. Although most-similar sampling achieves the best text generation quality by focusing on highly relevant samples, it overlooks instances with low product similarity but potential reference value, thus offering only limited gains in PBS. In contrast, our sampling strategy maintains both relevance and diversity, achieving the best overall balance between image and text generation performances.

\section{Instruction Templates}
\label{supp:instruction_template}
Figure \ref{fig:instruction_template} presents the instruction templates used in our framework. The ``\{\}" indicates the placeholder for textual input, while the special tokens $\texttt{<image\_ph>}$ and $\texttt{<text\_ph>}$ mark the insertion location for image and text preference token embeddings from multimodal preference extraction. These templates are randomly employed during training to enhance the model's ability to mitigate hallucinations, effectively constraining text generation to the predefined selling points and ensuring accurate, controllable output.

\section{Limitation}
\label{supp:limitation}
While this study introduces the first unified framework for personalized advertising image-text generation, several limitations remain. Existing autoregressive frameworks struggle with embedded visual text generation, typically producing poor results. Our work therefore focuses on establishing a reliable baseline for personalized advertisements without visual text, leaving text rendering for future work. Besides, the inference efficiency of our method cannot meet the requirements of real-time inference. Future efforts may incorporate visual text control modules and optimize model architecture to enable generation for industrial applications \cite{wang2025dualnet,wang2025adstereo,wang2025wisa,ling2025ragar,cao2025relactrl,ma2025lay2story,zhang2025u}.

\section{Social Impact}
\label{supp:social_impact}
\noindent\textbf{Practical impact.} The proposed method could serve as a heuristic tool to collaborate with human designers, helping them reduce the routine workload while complementing their expertise. By automatically producing image-text content aligned with user preferences, it significantly reduces manual design efforts. This technology enables e-commerce platforms to achieve more precise advertisement targeting, improving product exposure efficiency and increasing user click-through rates, thereby advancing intelligent digital marketing.

\noindent\textbf{Ethical impact.} The PAd1M dataset is constructed in strict compliance with legal requirements and platform regulations. All advertising content underwent a thorough review by the e-commerce platform to ensure no ethical concerns or sensitive information were raised. The dataset is solely for academic research and has passed a comprehensive legal review regarding copyright and usage scope, guaranteeing its ethical application within the research area without commercial infringement risks.

\clearpage

\end{document}